\documentclass[runningheads]{llncs}

\usepackage{eccv}
\usepackage{eccvabbrv}
\usepackage{afterpage}

\usepackage{graphicx}
\usepackage{booktabs}
\newcommand{\mname}{SCENT}

\usepackage{pifont}
\newcommand{\cmark}{\ding{51}}
\newcommand{\xmark}{\ding{55}}

\usepackage{xcolor}

\definecolor{obj_color}{HTML}{0792A5}
\definecolor{ctx_color}{HTML}{AF8900}
\definecolor{inf_color}{HTML}{389E34}
\definecolor{smell_cyan}{HTML}{77DDCC}
\definecolor{image_orange}{HTML}{F79256}

\usepackage{makecell}

\usepackage[accsupp]{axessibility}  

\usepackage{multirow}
\usepackage{chngcntr}
\usepackage{comment}

\usepackage{hyperref}
\usepackage{orcidlink}

\usepackage{multirow}
\usepackage{array}
\usepackage{siunitx}
\usepackage{booktabs}
\usepackage{calc}
\usepackage{bm}
\usepackage{graphicx}
\usepackage{subcaption}
\usepackage{pgfplots}
\pgfplotsset{compat=1.18}
\usepgfplotslibrary{groupplots}

\newcommand{\numbox}[1]{\makebox[2.5em][r]{#1}}
\newcommand{\deltasize}{\tiny} 
\newcommand{\gain}[1]{\textcolor{green!55!black}{\deltasize(+#1)}}
\newcommand{\loss}[1]{\textcolor{red!70!black}{\deltasize(-#1)}}
\newcommand{\eqdelta}{\textcolor{gray}{\deltasize(0.0)}}
\newcommand{\hidedelta}[1]{\textcolor{black!0}{\deltasize(#1)}} 
\newcommand{\tightbold}[1]{{\fontseries{b}\selectfont #1}}

\begin{document}

\title{What Images Cannot Say: Language-Guided Olfactory Representation Learning}

\titlerunning{What Images Cannot Say}

\author{Eleftherios Tsonis \and
Xi Wang \and
Vicky Kalogeiton}
\authorrunning{E.~Tsonis et al.}
\institute{LIX, École Polytechnique, IP Paris, CNRS\\[0.5em]
\email{\{firstname.lastname\}@polytechnique.edu}\\[0.5em]
\url{https://www.lix.polytechnique.fr/vista/projects/2026_scent_tsonis/}}

\maketitle

\begin{abstract}
Images tell us what a scene looks like, but rarely what it would feel like to be there. 
While recent datasets pair visual scenes with electronic-nose measurements, aligning smell signals with images remains challenging because many olfactory cues arise from contextual environmental factors that are not directly visible in pixels.
We introduce \mname{}, a multimodal framework that uses language guidance as a semantic bridge between vision and olfaction. Our approach leverages Vision-Language Models (VLMs) to generate scene descriptors capturing objects, environmental context, and plausible ambient smell cues suggested by the visual scene. These descriptors provide semantic guidance for learning olfactory representations.
We train a smell encoder that maps electronic-nose signals into a shared embedding space aligned with both visual and textual representations, and introduce a language-guided latent decomposition that separates object-specific odors from contextual environmental contributions.
Experiments on the New York Smells dataset demonstrate that \mname{} significantly improves cross-modal retrieval compared to vision-only baselines, achieving state-of-the-art performance on smell-to-image and smell-to-text retrieval tasks. In addition, our framework produces interpretable olfactory representations that enable the disentanglement of complex smell mixtures.
Our results reveal the importance of contextual semantic information for grounding olfactory perception in multimodal learning and pave the way for future research in this area.
\end{abstract}

\section{Introduction}
\label{sec:intro}
\begin{quote}
    \begin{flushright}
        \itshape \small{``A picture is worth a thousand words'' (att.\ Frederick R. Barnard)} \\
        \small{... but it rarely tells us what it \textbf{feels} like to be there..}
    \end{flushright}
\end{quote}

In real environments, perception extends beyond vision. 
A photograph of a busy street may evoke the smell of vehicle exhaust; a metro entrance may imply the metallic scent of ventilation air. 
Although such environmental cues are rarely visible directly, humans routinely infer them from visual context. Despite years of progress in computer vision, AI systems today primarily capture \emph{what a scene looks like}, while remaining largely unaware of what it might \emph{feel like to be there}.

Recent advances in Vision-Language Models (VLMs) suggest that this long-standing idea may finally be realized: modern systems can generate rich descriptions of visual scenes~\cite{qwen3technicalreport,li2023blip,lillava}, answer questions about images~\cite{Xiao_2024_CVPR,zhang2025videollama,fang2024mmbench}, and reason about complex environments through language~\cite{kimopenvla,black2024pi_0}. Their extensive world knowledge opens the possibility of inferring what exists in a scene, beyond what is visible.

\begin{figure}[t]
    \centering    \includegraphics[width=\linewidth]{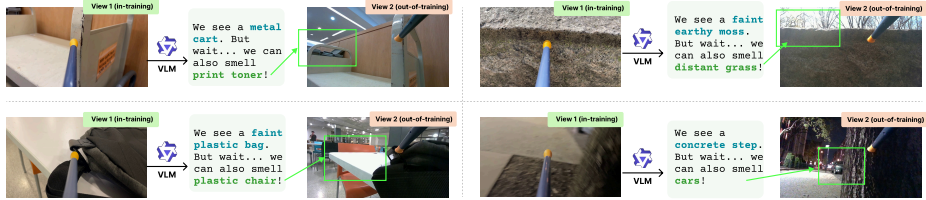}
    \caption{\textbf{Not everything we can smell is visible.} Given only View 1 (in-sample) during training, the true olfactory context may lie outside the field of view, in View 2 (out-of-sample). A VLM can bridge this gap by inferring plausible smells from semantic context alone. We use these language-derived signals as supervision to learn richer smell representations that go beyond what is directly seen.}
    \label{fig:toy_ex}
\end{figure}

Among sensory modalities, olfaction is particularly important.
Smell conveys rich semantic information that is difficult to infer from other modalities~\cite{stevenson2010,furizal2023future}, including food quality and safety \cite{tan2020applications, ali2020principles}, environmental conditions and airborne hazards, such as pollution \cite{brattoli2011odour} and toxic substances \cite{wilson2012review, chen2022gas}, as well as social cues related to health and emotional state \cite{kontaris2020behavioral, moein2020smell}.

Advances in electronic noses (e-noses)~\cite{wilson2012review} enable the capture of high dimensional sensor measurements~\cite{ozguroglu2025new,feng2025smellnet,erlangga2021electronic} that characterize the chemical composition of air. Mapping these raw signals to semantic representations is, however, fundamentally difficult. First, the image paired with each recording captures only a \emph{partial view} of the scene: smells diffuse freely and may originate from sources entirely outside the camera's field of view, so visual supervision alone cannot fully account for what was measured (Figure~\ref{fig:toy_ex}). Second, a smell recording reflects a mixture of contributions from multiple odor sources in the scene, making it hard to disentangle individual semantic factors. 

VLMs trained on large-scale data develop world knowledge that extends well beyond visual appearence, learning to reason about physical, acoustic, and semantic properties of scenes~\cite{huh2024platonic, yu2024kola, wang2025teaching}. This extends to olfaction, as Large Language Models (LLM) can classify odors, predict smell descriptors, and identify sources from natural language descriptions, demonstrating that smell associations are encoded through semantic context~\cite{makri2026benchmark}. Inherited from LLMs, modern VLMs also hold natural knowledge priors which are useful for our problem: given a scene image, a VLM can generate plausible olfactory descriptions that go beyond what the camera captures, directly compensating for the partial observability of visual supervision.

Building on this insight, we propose \textbf{\mname{}}: \textbf{S}emantic \textbf{C}ontext-aware \textbf{e}-\textbf{N}ose \textbf{T}ransformer, a framework that integrates e-nose measurements with visual and linguistic representations. 
Our approach first uses a VLM to generate structured textual descriptors of a scene, capturing objects, environmental context, and plausible ambient smell cues. 
These descriptions serve as a semantic bridge between vision and smell: they encode what a scene implies about its olfactory environment, extending supervision beyond what the camera alone can observe.
The descriptors are encoded and used as semantic anchors for representation learning. 
We train a smell encoder that maps e-nose signals into a shared embedding space aligned with both visual and textual representations, enabling multimodal representation of olfactory environments.

To further structure the learned representation, we introduce a latent decomposition that separates object-specific odor signals from contextual environmental contributions. 
This decomposition is guided by the text descriptions produced by the VLM and encourages the model to disentangle the components of complex olfactory mixtures.

We evaluate our approach on the New York Smells (NYS) dataset~\cite{ozguroglu2025new} and demonstrate that integrating language guidance significantly improves multimodal alignment compared to vision-only baselines. 
Our method achieves state-of-the-art performance on cross-modal retrieval tasks, including smell-to-image and smell-to-text retrieval, while also producing more interpretable olfactory representations.
In summary, our contributions are:
\begin{itemize}

\item We introduce the idea of using \textbf{language guidance as a semantic bridge} between visual scenes and olfactory signals, enabling the interpretation of smell measurements through contextual world knowledge.

\item We present \mname{}, a \textbf{multimodal framework for learning olfactory representations} that aligns e-nose, visual, and textual embeddings.

\item We propose a language-guided \textbf{decomposition of olfactory representations} that separates object-related odors from environmental contributions.

\end{itemize}

\section{Related Work}
\label{sec:related}
\paragraph{Cross-modal and multimodal supervision.}
Using one modality to supervise another has a long history, spanning early self-supervised approaches~\cite{de1993learning} and deep multimodal models~\cite{ngiam2011multimodal}, to more recent methods that exploit naturally co-occurring sensor data for audio-visual~\cite{owens2016visually,aytar2016soundnet} and visual-tactile~\cite{yuan2017connecting,yang2022touch,dou2024tactile} learning.
Models such as CLIP~\cite{radford2021learning} and ALIGN~\cite{Jia2021ScalingUV} learn joint image-text representations, while CLAP~\cite{CLAP2022} extends this paradigm to audio and text. Similar approaches have since been applied to video–text~\cite{Luo2021CLIP4ClipAE} and point cloud-text~\cite{Zhang2021PointCLIPPC} representation learning.
Works such as CLIP4VLA~\cite{Ruan2023AccommodatingAM}, VALOR~\cite{Chen2023VALORVO}, VAST~\cite{Chen2023VASTAV}, mPLUG-2~\cite{Xu2023mPLUG2AM}, UMT~\cite{liu2022umt}, InternVideo2~\cite{Wang2024InternVideo2SV}, and ConFu~\cite{koutoupis2026more} align three modalities at once.
For higher order settings, existing methods often designate one modality as an anchor and align the remaining modalities through pairwise contrastive objectives~\cite{mordacq2024adapt}. For example, ImageBind~\cite{Girdhar2023ImageBindOE} uses images as the central bridge, and LanguageBind~\cite{Zhu2023LanguageBindEV} uses text. OneLLM~\cite{han2024onellm} aligns all modalities to a frozen language model, and UniAlign~\cite{zhou2025unialign} encodes modalities in a unified mixture-of-experts architecture.
More recently, alternative formulations have been explored, including geometric approaches such as GRAM~\cite{cicchetti2025gram} and TRIANGLE~\cite{cicchettitriangle}, and information-theoretic objectives such as Symile~\cite{saporta2024contrasting} and CoMM~\cite{dufumier2025align}. %

\paragraph{Language as a semantic bridge across modalities}
Machine generated language provides a scalable alternative to manual annotation, enabling semantic supervision of modalities that are otherwise expensive to label. Early neural image captioning~\cite{vinyals2015show} established that vision and language can be jointly modelled to produce descriptive text from visual inputs. Later work leverages this ability as a supervisory signal across diverse modalities. ULIP-2~\cite{xue2024ulip} generates holistic language descriptions from rendered views of 3D shapes, enabling scalable tri-modal pre-training over point clouds, images, and text without any human annotation. LAVILA~\cite{zhao2023learning} similarly uses LLMs as narrators to densely annotate video, yielding supervision for contrastive video–text learning. Moving beyond vision, WineSensed/FEAST~\cite{bender2023learning} demonstrates that text and images can be aligned to human sensory perception of taste, establishing that sensory modalities can be grounded in shared multimodal representations. Building on these precedents, we use VLM-generated descriptors to inject world knowledge and contextual cues into olfactory representation learning.

\paragraph{Machine olfaction and olfactory representations.}
Machine olfaction has historically focused on constrained settings using specialized or laboratory-grade sensing~\cite{Debnath2020, Mueller2019, feng2025smellnet, rodriguez2009electronic, wijaya2021dwtlstm, vergara2012chemical}, enabling applications such as scent design~\cite{sanchez2019machine}, disease detection~\cite{fundurulic2023advances, ghazaly2023assessment}, and security~\cite{torres2020improving}. At the molecular level, prior work has used psychophysical data and graph-based models to predict perceptual attributes~\cite{Keller2017, dravnieks1985atlas} and to define a principal odor map (POM)~\cite{lee2023principal}, while mixture and similarity studies remain limited~\cite{Snitz2013,Ravia2020}. Recent findings from neuroscience emphasize the importance of studying olfaction under natural concentration ranges~\cite{Wachowiak2025}. Concurrently, recent work examines how well large language models can reason about smell~\cite{makri2026benchmark}.

Moving beyond lab-based limitations, the New York Smells (NYS) dataset \cite{ozguroglu2025new} is a large-scale multimodal benchmark designed for "in-the-wild" smell perception. It comprises 7,000 image-olfactory pairs, collected across diverse urban environments ranging from indoor libraries to outdoor parks. NYS frames olfactory learning as direct cross-modal alignment between visual imagery and e-nose signals. Our approach differs from NYS by using language as a semantic bridge to inject \emph{prior world knowledge} into smell representation learning, capturing olfactory cues that visual appearence alone cannot express. To our knowledge, no existing work combines \emph{olfactory sensor signals} with \emph{vision} and \emph{text} together.

\section{Method}
\label{sec:method}
We introduce \textbf{\mname{}}: \textbf{S}emantic \textbf{C}ontext-aware \textbf{e}-\textbf{N}ose \textbf{T}ransformer, a framework for learning olfactory representations from e-nose measurements, guided by vision and language. Our key insight is that an olfactory signal $\mathbf{X} \in \mathbb{R}^{C \times T}$ (a $C$-channel e-nose recording over $T$ time steps) reflects odor contributions from the \emph{entire} scene, yet paired visual observations capture only a subset of these factors (Figure~\ref{fig:toy_ex}). In other words, smell information is only \textbf{partially observable} from visual input.
To bridge this gap, we leverage the strong world priors of Vision-Language Models (VLMs)~\cite{qwen3technicalreport,lillava}, which can infer plausible olfactory cues from visual context.

\begin{figure}[t]
    \centering
    \includegraphics[width=\linewidth]{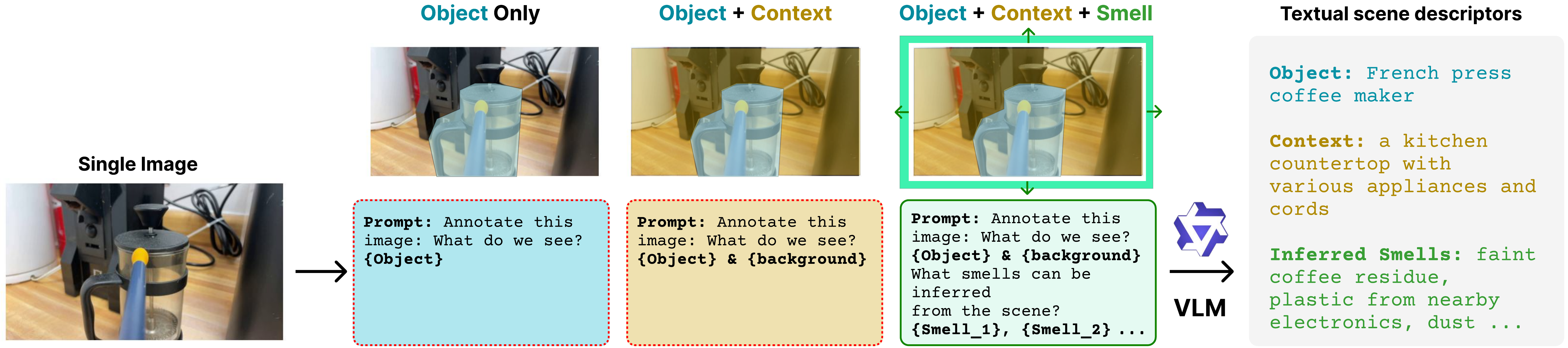}
    \caption{\textbf{Textual scene descriptors via VLM inference.} Given an image, a pretrained VLM generates object, contextual, and inferred smell descriptors, which serve as language guidance for olfactory representation learning.}
    \label{fig:vlm}
\end{figure}

\mname{} therefore combines three stages: (1) VLM-based semantic scene augmentation (Figure~\ref{fig:vlm}, Section~\ref{sub:method_stage1}); (2) multimodal olfactory representation learning (Figure~\ref{fig:both_images}(a), Section~\ref{sub:method_stage2}); and (3) latent smell disentanglement that separates object and contextual odor components (Figure~\ref{fig:both_images}(b), Section~\ref{sub:method_stage3}). 

\subsection{VLM-based Semantic Scene Augmentation}
\label{sub:method_stage1}

The NYS dataset provides images $I$ and corresponding smell measurements $\mathbf{X}$. To expose the environmental factors that shape olfactory perception, we query a pretrained VLM through a structured prompt, eliciting its prior world knowledge at three semantic levels.
Given an image $I$, the VLM produces a set of language descriptors: 
\[
T(I) = \{t_{\text{obj}}, t_{\text{ctx}}, t_{\text{smell}}^{(1)}, \ldots, t_{\text{smell}}^{(K)}\} \quad,
\]
where $t_{\text{obj}}$ denotes a textual description of the primary object in the scene, $t_{\text{ctx}}$ describes the surrounding environmental context, and $\{t_{\text{smell}}^{(k)}\}_{k=1}^K$ are textual descriptors referring to plausible ambient smell cues suggested by the scene context. 
These descriptors provide semantic cues that may relate to the measured olfactory signal but are not directly encoded in the image alone. As shown in Figure~\ref{fig:vlm}, we explore several prompting schemes and adopt the rightmost one, which explicitly encourages the VLM to infer potential olfactory elements.

These descriptors are combined and prompted into a textual description and encoded using the frozen CLIP~\cite{radford2021learning} text encoder: ${z^T = f_T(\text{prompt}(T)) \in \mathbb{R}^{512}}$.
The resulting representation $z^T$ serves as language supervision in the multimodal alignment stage (Section~\ref{sub:method_stage3}), helping compensate for information missing from visual observations.

\subsection{Learning Multi-modal Olfactory Representations}
\label{sub:method_stage2}

\begin{figure}[tbp]
     \centering
     \begin{subfigure}[b]{0.39\textwidth}
         \centering
         \includegraphics[width=\textwidth]{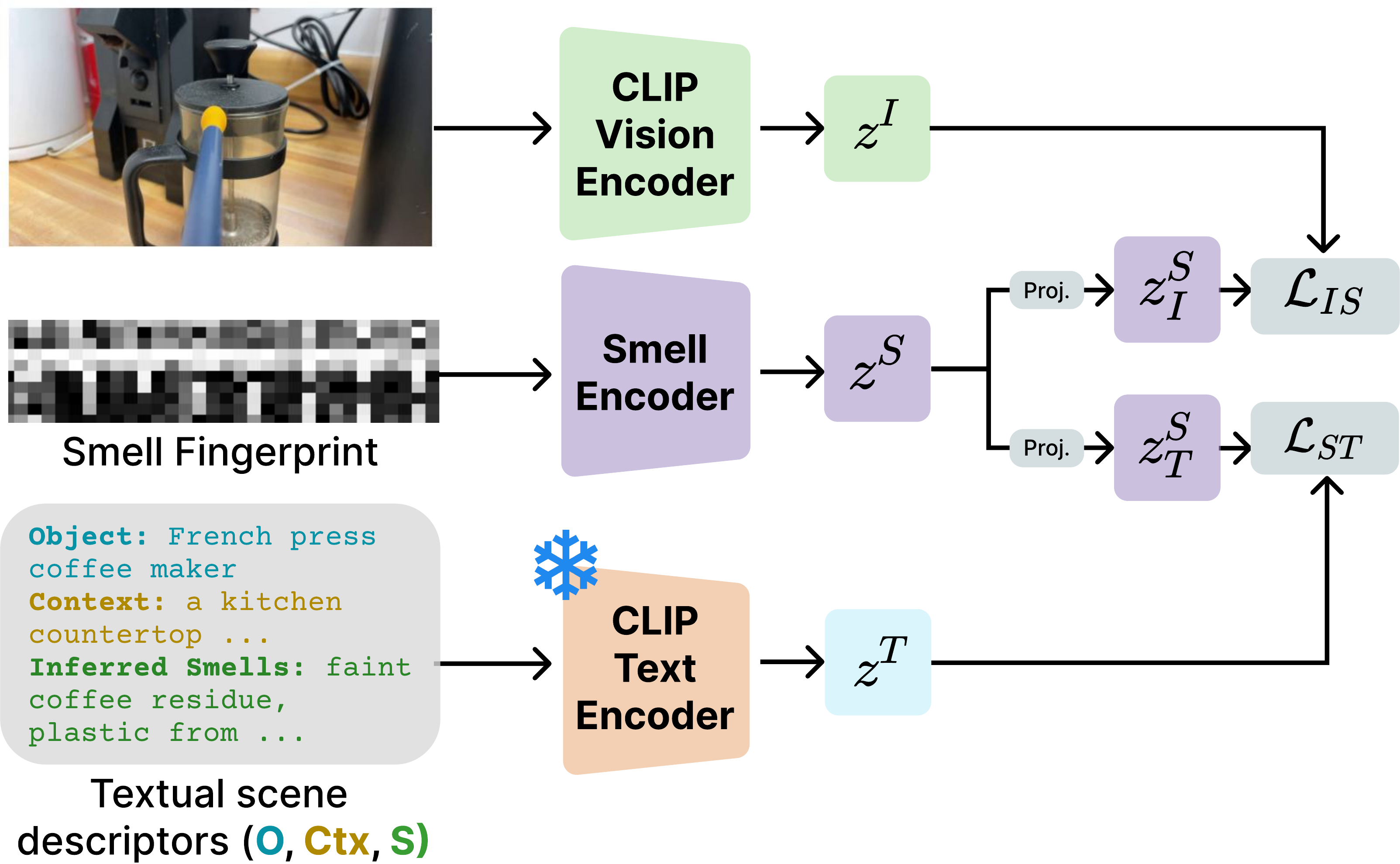}
         \caption{Smell representation learning.}
         \label{fig:left}
     \end{subfigure}
     \hfill
     \begin{subfigure}[b]{0.60\textwidth}
         \centering
         \includegraphics[width=\textwidth]{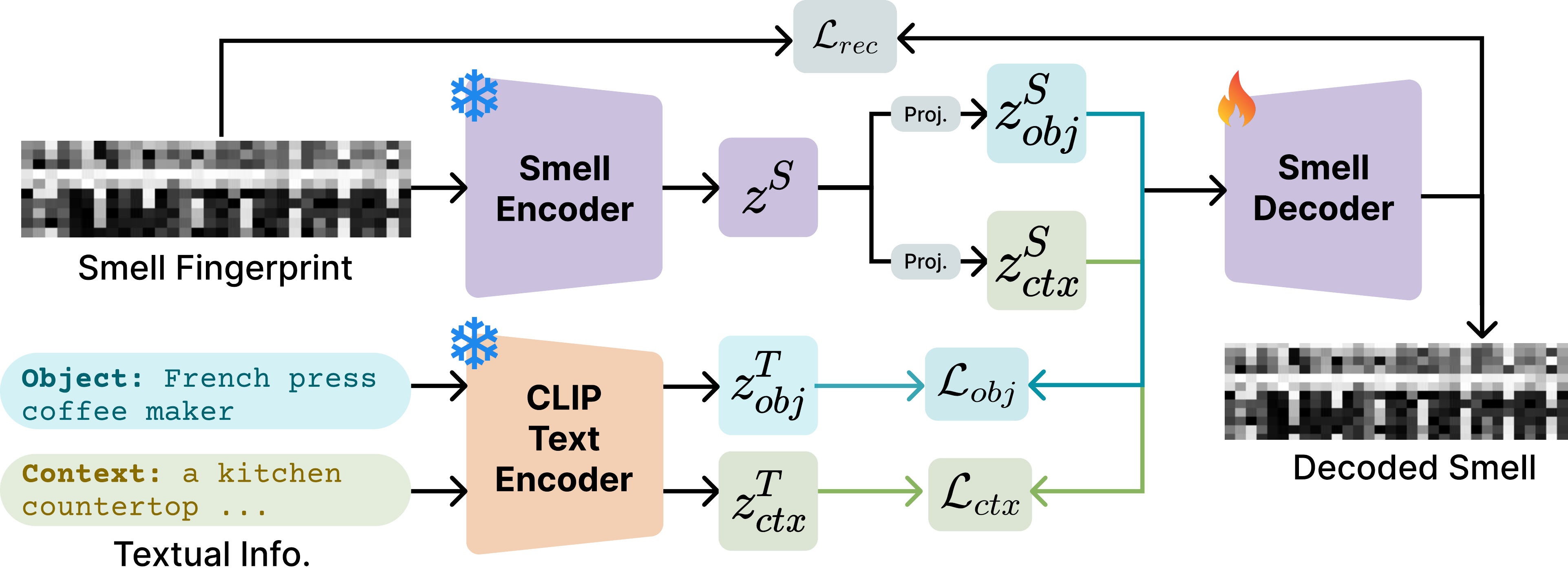}
         \caption{Latent smell disentanglement.}
         \label{fig:right}
     \end{subfigure}
     \caption{\textbf{Overall framework for multimodal olfactory representation learning and disentanglement.} \textbf{(a)} Smell Representation Learning: We train a Smell Encoder to align smell fingerprints with CLIP visual ($z^I$) and textual ($z^T$) embeddings. The dual projection heads yield $z_{I}^S$ and $z_{T}^S$. \textbf{(b)} Latent Smell Disentanglement: Leveraging the pretrained encoder, we disentangle the smell representation into object ($z_{\text{obj}}^S$) and context ($z_{\text{ctx}}^S$) components. These components are aligned with their respective linguistic descriptors via $\mathcal{L}_{\text{obj}}$ and $\mathcal{L}_{\text{ctx}}$.
     }
     \label{fig:both_images}
\end{figure}

With visual observations $I$ and augmented smell descriptors $T(I)$ available, we learn an olfactory representation aligned with both visual and textual modalities.

\paragraph{Olfactory Encoder.}
The smell signal $\mathbf{X}$ is processed using a Transformer-based encoder $f_S$ that models temporal dependencies across sensor measurements, yielding $z^S = f_S(\mathbf{X})$. This embedding captures the global structure of the olfactory signal. To enable alignment with visual and textual modalities, we apply two modality-specific projection heads:
\[
z^S_I = \phi_I(z^S) \in \mathbb{R}^{512}, \qquad
z^S_T = \phi_T(z^S) \in \mathbb{R}^{512},
\]
which map the olfactory representation toward the visual and textual embedding spaces, respectively (Figure~\ref{fig:left}).

\paragraph{Visual and Textual Encoders.}
We obtain visual embeddings $z^I$ using CLIP's~\cite{radford2021learning} image encoder, and textual embeddings $z^T$ using CLIP's text encoder, as described in Section~\ref{sub:method_stage1}. We have: 

\[
z^I = f_I(I) \in \mathbb{R}^{512}, \qquad
z^T = f_T\left(\text{prompt}(t_{\text{obj}}, t_{\text{ctx}}, t_{\text{smell}}^{(1)}, \dots, t_{\text{smell}}^{(K)})\right) \in \mathbb{R}^{512}.
\]

While the text encoder remains frozen to preserve semantic structure, the image encoder is fine-tuned to adapt its features towards olfactory-relevant cues. %

\paragraph{Multimodal Alignment.}
We train the smell encoder by aligning the projected smell representations with their corresponding visual and textual embeddings using symmetric InfoNCE losses \cite{oord2018representation,radford2021learning}. Let $\mathrm{sim}(\cdot,\cdot)$ denote cosine similarity. For a batch of $N$ samples we compute
\[
\resizebox{0.9\linewidth}{!}{$
\begin{aligned}
\mathcal{L}_{IS} = -\frac{1}{2N} \sum_{i=1}^{N} \left(
\log \frac{\exp(\mathrm{sim}(z^S_{I,i}, z^I_i)/\tau)}
{\sum_{j=1}^{N} \exp(\mathrm{sim}(z^S_{I,i}, z^I_j)/\tau)}
+
\log \frac{\exp(\mathrm{sim}(z^S_{I,i}, z^I_i)/\tau)}
{\sum_{j=1}^{N} \exp(\mathrm{sim}(z^S_{I,j}, z^I_i)/\tau)}
\right)
\end{aligned}
$}
\]
where $\tau$ is a learnable temperature parameter. $\mathcal{L}_{ST}$ is computed analogously using $(z^S_T, z^T)$.
The overall representation learning objective is
\begin{equation}
\label{eq:loss_total}
\mathcal{L}_{\text{total}}
=
\lambda_{IS}\mathcal{L}_{IS}
+
\lambda_{ST}\mathcal{L}_{ST} \quad.
\end{equation}

This stage produces an olfactory embedding that is jointly structured by the smell measurements, the visual scene, and the augmented language descriptors generated by the VLM.

\subsection{Latent Smell Disentanglement}
\label{sub:method_stage3}

The representation $z^S$ learned in the previous stage captures the overall olfactory signal present in the measurement. 
However, smell recordings often contain a mixture of sources, including odors emitted from the target object as well as background environmental factors.
To disentangle these contributions, we introduce a latent decomposition of the olfactory representation into object-specific and background-related components. This step is illustrated in Figure~\ref{fig:right}. 

\paragraph{Latent Decomposition.}
Our goal is to decompose the learned embedding into two components:
\[
z^S_{\text{obj}} = \phi_{\text{obj}}(z^S_T) 
\qquad \text{and} \qquad
z^S_{\text{ctx}} = \phi_{\text{ctx}}(z^S_T) \quad,
\]
where $z^S_{\text{obj}}$ and $z^S_{\text{ctx}}$ \emph{exclusively} represent object-related odor factors and the remaining contextual environmental contributions, respectively.

\paragraph{Semantic Alignment.} The decomposition is enabled by the practical compositionality of language descriptors, i.e. we have separate text encodings for an `object' and the `context' of the scene it is located in. During training, each decoded smell component is aligned with its corresponding textual descriptor obtained in Section~\ref{sub:method_stage1} to provide explicit supervision: the object component $z^S_{\text{obj}}$ is aligned with the object descriptor $t_{\text{obj}}$, while the contextual component $z^S_{\text{ctx}}$ is aligned with the context descriptor $t_{\text{ctx}}$.

Denoting $z^T_{\text{obj}}$ and $z^T_{\text{ctx}}$ the corresponding CLIP text embeddings. We enforce this decomposition through contrastive losses. Let $\mathrm{sim}(\cdot,\cdot)$ denote cosine similarity. For a batch of $N$ samples, we compute:

\[
\begin{aligned}
\mathcal{L}_{obj}
&=
-\frac{1}{N}\sum_{i=1}^{N}
\log
\frac{\exp(\mathrm{sim}(z^S_{obj,i},z^T_{obj,i})/\tau)}
{\sum_{j=1}^{N}\exp(\mathrm{sim}(z^S_{obj,i},z^T_{obj,j})/\tau)}
\\[6pt]
\mathcal{L}_{ctx}
&=
-\frac{1}{N}\sum_{i=1}^{N}
\log
\frac{\exp(\mathrm{sim}(z^S_{ctx,i},z^T_{ctx,i})/\tau)}
{\sum_{j=1}^{N}\exp(\mathrm{sim}(z^S_{ctx,i},z^T_{ctx,j})/\tau)} .
\end{aligned}
\]

\paragraph{Reconstruction Constraint.}
To ensure that the decomposition preserves the information contained in the original smell signal without representation collapse, we also implement a reconstruction objective. 
A decoder $d(\cdot)$ predicts the smell signal from the concatenated latents:

\[
\mathcal{L}_{\text{rec}} = \|\mathbf{X} - {d([z^S_{\text{obj}}, z^S_{\text{ctx}}])}\|^2 \quad .
\]

The disentanglement stage is therefore a weighted combination of these losses:
\begin{equation}
\label{eq:loss_dise}
\mathcal{L}_{\text{dis}} =
\mathcal{L}_{\text{obj}} +
\mathcal{L}_{\text{ctx}} +
\lambda_{\text{rec}} \mathcal{L}_{\text{rec}} \quad.
\end{equation}
\noindent $\lambda_{rec}$ is a weighting factor for the reconstruction loss, which acts as a regularizer to prevent embedding collapse. 
This decomposition encourages \mname{} to separate object-specific odor signals from contextual environmental contributions while remaining consistent with the underlying sensor measurements.

\section{Experiments}
\label{sec:experiments}
\subsection{Implementation details and experiment settings}

\paragraph{Dataset.}
\noindent
We evaluate \mname{}
on the \textbf{New York Smells (NYS)} dataset \cite{ozguroglu2025new}, the largest multimodal benchmark for in-the-wild olfactory perception. NYS contains 7,000 paired image-smell samples across 3,500 distinct object categories. We use 5,996 training and 936 validation samples; all baselines and ablations share this split. 

\paragraph{Implementation Details.}
Each olfactory sample comprises a 32-channel resistance signal from an e-nose array, consisting of a baseline recording ($\mathbf{B} \in \mathbb{R}^{C \times T}$) of ambient air and a sample recording ($\mathbf{S} \in \mathbb{R}^{C \times T}$) taken near the object, where $C = 32$ is the number of sensor channels and $T$ denotes the temporal dimension (typically 14 timesteps). Following \cite{ozguroglu2025new}, we use a unified olfactory input $\mathbf{X}$ formed by the temporal concatenation of the two signals: \mbox{$\mathbf{X} = [\mathbf{B}; \mathbf{S}] \in \mathbb{R}^{C \times 2T}$}.
Our olfactory encoder $f_S$ is a Transformer with $6$ layers and $8$ attention heads. The input signal $\mathbf{X}$
is projected to a $448$ dimensional embedding and augmented with sinusoidal positional encodings, the resulting olfactory representation $z^S$ is then mapped to the $512$-dimensional CLIP visual and textual spaces using two MLP projection heads. For semantic augmentation, we use Qwen3VL-30B \cite{qwen3technicalreport} to generate descriptors $\{t_{\text{obj}}, t_{\text{ctx}}, t_{\text{smell}}\}$. We employ CLIP ViT-B/16 \cite{radford2021learning} as our image-language encoder backbone. During training, the text encoder remains frozen, while the vision encoder is trainable, unless stated otherwise in specific ablation studies. Additional architectural and training details are provided in the supplementary material.

\paragraph{Evaluation Metrics and Retrieval Tasks.}
We evaluate \mname{} across two primary categories, single-modality and joint-modality retrieval, to assess the alignment of olfactory features with individual and combined semantic spaces. 
We report standard retrieval metrics: Recall@$k$ (\%) (R@$k$ for $k \in \{1, 5, 10, 20\}$).

\noindent\textbf{Single-modality Retrieval (S2I, S2T):} These tasks evaluate the model's ability to align olfactory signals with a single target modality. In Smell-to-Image (S2I) and Smell-to-Text (S2T), a query smell signal is used to rank and retrieve candidates from a set of images or textual descriptors, respectively, based on cross-modal similarity.

\noindent\textbf{Joint-modality Retrieval (S2IT):} This task evaluates the model’s performance in a fully multimodal setting. The query is a singular olfactory signal, and the search database consists of (Image, Text) pairs. For each candidate pair in the gallery, the model must compute a joint similarity score that accounts for both the visual and textual components. We compute this score via latent-level fusion of the dual olfactory projections; full details and a comparison with an alternative fusion strategy are provided in the supplementary material.

\paragraph{Baselines.} 
As the current state-of-the-art on the NYS dataset~\cite{ozguroglu2025new} is a vision-only model, it serves as our primary comparison for the S2I task. However, since this baseline lacks a native textual projection, it cannot be directly applied to tasks involving language. Moreover, the original work does not provide publicly available code or pretrained weights, requiring us to reproduce the model based on the descriptions in the paper. To establish a rigorous baseline for text-based tasks, we reproduce the NYS model and implement NYS (adapted) for multi-modal retrieval tasks, which utilizes a two-stage ``Image Bridge'' (IB) protocol:
\begin{itemize}
\item \textbf{S2T Retrieval:} The model first retrieves the top-1 most visually similar image to the query smell using its olfactory-visual head. This ``bridge'' image is then encoded via a frozen CLIP ViT-B/16 \cite{radford2021learning} model and used to rank textual candidates in the CLIP latent space.
\item \textbf{Joint Retrieval (S2IT):} For S2IT, a composite score is formed by summing the smell-image similarity (from the NYS model) with the pre-existing image-text alignment score (from CLIP) for each candidate pair.
\end{itemize}

\subsection{The Necessity of Language Guidance}
\label{sub:exp_ablation}
{

\begin{table}[t]
\centering
\renewcommand{\arraystretch}{1.3}
\setlength{\tabcolsep}{5pt}
\caption{\textbf{Impact of semantics
on olfactory alignment.} We evaluate the retrieval performance of our olfactory encoder when aligned with frozen CLIP visual and textual spaces. Results indicate that simple object labels ($\textbf{\textcolor{obj_color}{O}}$) provide insufficient supervision, whereas incorporating 
context ($\textbf{\textcolor{ctx_color}{Ctx}}$), and inferred ambient smells ($\textbf{\textcolor{inf_color}{I}}$) yields the 
better cross-modal alignment across both single and joint retrieval tasks. Best results in \textbf{bold}.}
\resizebox{.9\textwidth}{!}{%
\begin{tabular}{l ccc lll} 
\toprule
\textbf{Align. Strategy} & \multicolumn{3}{c}{\textbf{Textual Info}} & \multicolumn{3}{c}{\textbf{Smell Retrieval}} \\
\cmidrule(lr){2-4} \cmidrule(lr){5-7}
& $\textbf{\textcolor{obj_color}{O}}$ & $\textbf{\textcolor{ctx_color}{Ctx}}$ & $\textbf{\textcolor{inf_color}{S}}$ & \numbox{R@5} & \numbox{R@10} & \numbox{R@20} \\
\midrule
Image (S2I) & & & & \numbox{12.1}\hidedelta{0.0} & \numbox{18.3}\hidedelta{0.0} & \numbox{29.2}\hidedelta{0.0} \\
\midrule
Text (S2T) & \checkmark & & & \numbox{8.8} \loss{3.3} & \numbox{14.4} \loss{3.9} & \numbox{24.9} \loss{4.3} \\
& \checkmark & \checkmark & & \numbox{10.8} \loss{1.3} & \numbox{18.3} \eqdelta & \numbox{28.1} \loss{1.1} \\
& \checkmark & \checkmark & \checkmark & \numbox{12.4} \gain{0.3} & \numbox{19.2} \gain{0.9} & \numbox{29.0} \loss{0.2} \\
\midrule
\textbf{Image \& Text (S2IT)} & \checkmark & \checkmark & \checkmark & \numbox{\tightbold{14.3}} \gain{2.2} & \numbox{\tightbold{22.5}} \gain{4.2} & \numbox{\tightbold{35.1}} \gain{5.9} \\
\bottomrule
\end{tabular}
}
\label{tab:exp1_results}
\end{table}
}
We first motivate \mname{} by evaluating the impact of different supervision levels while keeping visual and textual backbones frozen (Table~\ref{tab:exp1_results}).
Aligning smell solely with images (S2I) achieves a baseline R@5 of 12.1.
When aligning strictly with text (S2T), we observe that performance is highly dependent on the semantic granularity of the descriptors. Using only isolated object labels ($\textbf{\textcolor{obj_color}{O}}$) results in poor alignment (8.8 R@5), as these labels fail to capture the environmental factors present in the sensor data. However, as we incrementally add contextual descriptors ($\textbf{\textcolor{ctx_color}{Ctx}}$) and VLM-inferred ambient smells ($\textbf{\textcolor{inf_color}{S}}$), performance improves significantly, eventually surpassing the image-only baseline. The most robust representations are formed through the joint \textbf{Image\&Text} strategy. By aligning the smell encoder with both modalities simultaneously using our semantic-rich template ($\textbf{\textcolor{obj_color}{O}} + \textbf{\textcolor{ctx_color}{Ctx}} + \textbf{\textcolor{inf_color}{S}}$), we achieve 14.3 R@5. This confirms that language guidance offers a critical semantic bridge
that visual features alone cannot provide.

\subsection{Comparison to State-of-the-Art}
In Table~\ref{tab:exp2_results}, we compare \mname{} against the state-of-the-art NYS benchmark.
\paragraph{S2I Retrieval.} 
We observe that language guidance
improves performance even on purely visual tasks. In the S2I setting, \mname{} 
outperforms the reproduced NYS baseline (R@5 from 20.0 to 22.1) even 
only with object labels $\textbf{\textcolor{obj_color}{O}}$. This confirms that the semantic structure of language helps the olfactory encoder extract more discriminative features than vision alone. As contextual ($\textbf{\textcolor{ctx_color}{Ctx}}$) and inferred smell ($\textbf{\textcolor{inf_color}{S}}$) descriptors are added, S2I performance peaks at 23.0 R@5, indicating that a more complete semantic description of the scene complements the visual signal and improves the olfactory representation.

\paragraph{S2T and S2IT Retrieval.} On tasks involving language, the advantage of our proposed architecture is even more apparent. The NYS (adapted) baseline struggles with the semantic gap, as its ``Image Bridge'' is limited by what is explicitly visible. In contrast, \mname{} leverages its native textual projection head to achieve 11.9 R@5 on S2T retrieval, a 47\% relative improvement over the best adapted baseline. Furthermore, our joint-modality retrieval (S2IT) results show that \mname{} effectively fuses visual and textual cues, outperforming the baseline by over 7\% in R@5.

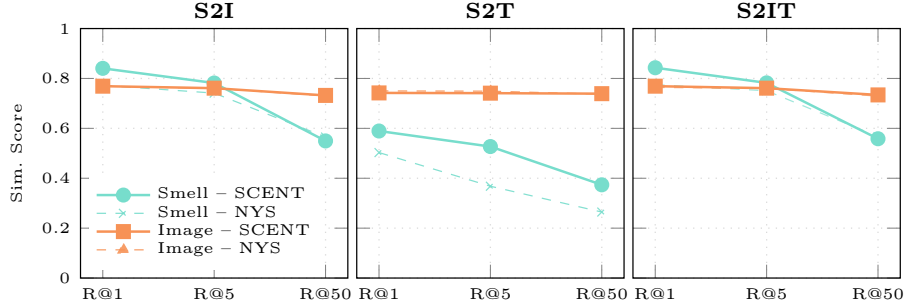
\begin{figure}[t]
\centering
\resizebox{\linewidth}{!}{\begin{tikzpicture}
\begin{groupplot}[
    group style={
        group size=3 by 1,
        horizontal sep=0.1cm,
        ylabels at=edge left,
        yticklabels at=edge left,
    },
    scale only axis,
    width=3cm, height=2.8cm,
    symbolic x coords={R@1, R@5, R@50}, xtick=data,
    ymin=0, ymax=1.0, ytick={0,0.2,0.4,0.6,0.8,1.0},
    tick label style={font=\tiny},
    label style={font=\tiny},
    title style={font=\scriptsize\bfseries, yshift=-1.5ex},
    grid=both, grid style={dotted, opacity=0.6},
]
\nextgroupplot[
    title={S2I},
    ylabel={Sim.\ Score},
    legend pos=south west,
    legend style={font=\tiny, draw=none, inner sep=2pt, row sep=-3pt,
                  legend cell align={left}},
]
\addplot[smell_cyan, solid, thick, mark=*]
    coordinates {(R@1,0.8405)(R@5,0.7817)(R@50,0.5495)};
\addlegendentry{Smell -- \mname{}}
\addplot[smell_cyan, dashed, opacity=0.9, mark=x]
    coordinates {(R@1,0.7727)(R@5,0.7412)(R@50,0.5652)};
\addlegendentry{Smell -- NYS}
\addplot[image_orange, solid, thick, mark=square*]
    coordinates {(R@1,0.769)(R@5,0.761)(R@50,0.732)};
\addlegendentry{Image -- \mname{}}
\addplot[image_orange, dashed, opacity=0.9, mark=triangle*]
    coordinates {(R@1,0.769)(R@5,0.760)(R@50,0.730)};
\addlegendentry{Image -- NYS}

\nextgroupplot[title={S2T}]
\addplot[smell_cyan, solid, thick, mark=*]
    coordinates {(R@1,0.5893)(R@5,0.5270)(R@50,0.3737)};
\addplot[smell_cyan, dashed, opacity=0.9, mark=x]
    coordinates {(R@1,0.5029)(R@5,0.3680)(R@50,0.2645)};
\addplot[image_orange, solid, thick, mark=square*]
    coordinates {(R@1,0.742)(R@5,0.741)(R@50,0.739)};
\addplot[image_orange, dashed, opacity=0.9, mark=triangle*]
    coordinates {(R@1,0.750)(R@5,0.749)(R@50,0.737)};

\nextgroupplot[title={S2IT}]
\addplot[smell_cyan, solid, thick, mark=*]
    coordinates {(R@1,0.8429)(R@5,0.7822)(R@50,0.5583)};
\addplot[smell_cyan, dashed, opacity=0.7, mark=x]
    coordinates {(R@1,0.7754)(R@5,0.7506)(R@50,0.5668)};
\addplot[image_orange, solid, thick, mark=square*]
    coordinates {(R@1,0.769)(R@5,0.761)(R@50,0.734)};
\addplot[image_orange, dashed, opacity=0.7, mark=triangle*]
    coordinates {(R@1,0.764)(R@5,0.760)(R@50,0.729)};
\end{groupplot}
\end{tikzpicture}}
\caption{Mean cosine similarity between the ground-truth and top-$k$ retrieved items, in frozen CLIP image space (orange) and learned olfactory space (teal), for $k\in\{1,5,50\}$. Solid lines: \mname{}; dashed: NYS.}
\label{fig:similarity_scores}
\end{figure}

\paragraph{Visual vs.\ Olfactory Discriminativity.}
Figure~\ref{fig:similarity_scores} plots the mean cosine similarity between the ground-truth item and the top-$k$ retrieved items, measured in frozen CLIP image space (orange) and in the learned olfactory space (teal), for $k\in\{1,5,50\}$.
Image features are weakly discriminative: scores remain nearly flat across retrieval ranks, failing to separate good retrievals (R@1) from bad ones (R@50).
In contrast, olfactory features drop sharply with retrieval rank; \mname{}'s smell curve (solid) decays faster than NYS's (dashed), confirming both that smell is the necessary modality for this task and that \mname{} learns a more discriminative representation.
Notably, NYS achieves similar S2T visual similarity to \mname{} yet 32\% lower R@5, showing that visual proximity does not imply correct olfactory match.

\begin{figure}[t]
    \centering
    \includegraphics[width=\linewidth]{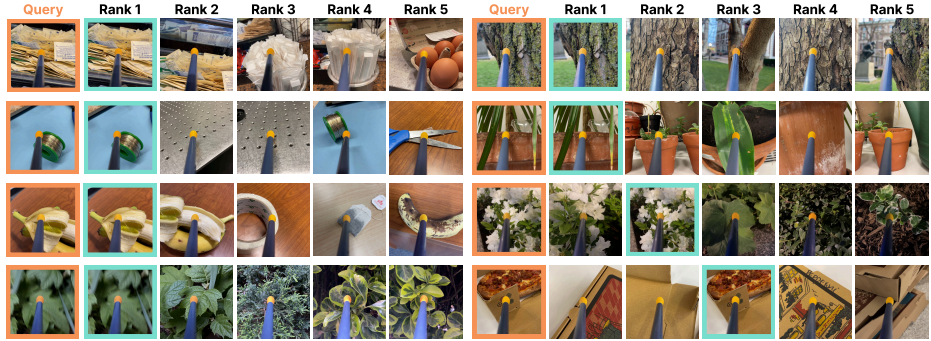}
    \caption{\textbf{Qualitative results for S2IT retrieval.} For a given \emph{olfactory} query, we show the ground-truth image in the \textbf{\textcolor{image_orange}{Query}} column and the top-5 images from the retrieved (Image, Text) pairs. The \textbf{\textcolor{smell_cyan}{Correct}} retrieval is highlighted. \mname{} demonstrates robust cross-modal alignment across diverse categories, successfully retrieving the correct match even when the visual appearance varies significantly from the query (e.g., the pizza box in Rank 1 vs. the open pizza in query).}
    \label{fig:qualitative}
\end{figure}

\paragraph{Qualitative Analysis.} We visualize the S2IT retrieval performance in Figure~\ref{fig:qualitative}. Each row corresponds to one example and the columns illustrate the top-5 images from the retrieved (Image,Text) pairs. Despite the inherent difficulty of the olfactory-to-visual mapping, \mname{} consistently ranks the correct semantic category within the top-5 results. Notably, the model handles fine-grained differences (e.g., distinguishing between different types of foliage or tree barks) and identifies objects regardless of their visual state, such as the peeled banana or the sliced pizza, by leveraging the underlying olfactory cue.

\begin{table}[t]
\centering
\renewcommand{\arraystretch}{1.2}
\caption{\textbf{State-of-the-art comparison on the NYS dataset.} We compare \mname{} against the vision-only NYS baseline~\cite{ozguroglu2025new} across single-modality (S2I, S2T) and joint-modality (S2IT) tasks. To enable a comparison on language-based tasks, we implement NYS (adapted) using an "Image Bridge" protocol. Our method consistently outperforms the adapted baseline, demonstrating that native three-stream alignment with inferred semantic cues captures nuanced olfactory information that is unobservable through visual features alone.
}
\setlength{\tabcolsep}{3pt}
\resizebox{\textwidth}{!}{%
\begin{small}
\begin{tabular}{l ccc rrrr rrrr rrrr}
\toprule
\multirow{2}{*}{\textbf{Method}} & \multicolumn{3}{c}{\textbf{Textual Info}} & \multicolumn{4}{c}{\textbf{S2I}} & \multicolumn{4}{c}{\textbf{S2T}} & \multicolumn{4}{c}{\textbf{S2IT}} \\
\cmidrule(lr){2-4} \cmidrule(lr){5-8} \cmidrule(lr){9-12} \cmidrule(lr){13-16}
& $\textbf{\textcolor{obj_color}{O}}$ & $\textbf{\textcolor{ctx_color}{Ctx}}$ & $\textbf{\textcolor{inf_color}{S}}$ & R@1 & R@5 & R@10 & R@20 & R@1 & R@5 & R@10 & R@20 & R@1 & R@5 & R@10 & R@20 \\
\midrule

NYS$^{*}$ \cite{ozguroglu2025new} & & & & --- & 16.5 & 29.6 & 43.1 & \multicolumn{4}{c}{---} & \multicolumn{4}{c}{---} \\
NYS$^{\dagger}$ & & & & 5.4 & 20.0 & 29.9 & 42.0 & \multicolumn{4}{c}{---} & \multicolumn{4}{c}{---} \\
\midrule

\multirow{3}{*}{NYS$^{\dagger}$ (adapted)} 
& \checkmark & & & \multicolumn{4}{c}{---} & 1.7 & 6.2 & 10.4 & 14.4 & 3.9 & 15.6 & 25.8 & 39.1 \\
& \checkmark & \checkmark & & \multicolumn{4}{c}{---} & 2.5 & 8.7 & 12.7 & 18.3 & 3.9 & 15.9 & 25.2 & 38.7 \\
& \checkmark & \checkmark & \checkmark & \multicolumn{4}{c}{---} & 2.6 & 8.1 & 12.9 & 19.2 & 4.9 & 18.5 & 27.5 & 40.7 \\
\midrule

\multirow{3}{*}{Ours} 
& \checkmark & & & 5.6 & 22.1 & 32.7 & 42.4 & 1.6 & 8.0 & 13.8 & 21.5 & 5.9 & 22.0 & 32.4 & 42.6 \\
& \checkmark & \checkmark & & 5.9 & 21.8 & 32.4 & \tightbold{45.2} & 2.8 & 10.7 & 17.3 & 28.1 & 6.1 & 21.3 & 32.5 & \tightbold{45.7} \\
& \checkmark & \checkmark & \checkmark & \tightbold{6.0} & \tightbold{23.0} & \tightbold{33.5} & 43.6 & \tightbold{3.3} & \tightbold{11.9} & \tightbold{19.8} & \tightbold{29.2} & \tightbold{6.8} & \tightbold{23.3} & \tightbold{32.7} & 42.5 \\
\bottomrule
\end{tabular}
\end{small}
}
\vspace{-8pt}
\begin{flushleft}
\scriptsize
$^{*}$ Results reported in the original NYS paper \cite{ozguroglu2025new}. \\
$^{\dagger}$ Results reproduced.
\end{flushleft}
\label{tab:exp2_results}
\end{table}

\subsection{Ablation Studies}

\paragraph{Impact of Semantic Granularity.}
We first evaluate how the complexity of the textual descriptors impacts olfactory alignment. As shown in Table~\ref{tab:exp2_results}, we observe a consistent performance gain across all tasks as the supervision transitions from simple object labels ($\textbf{\textcolor{obj_color}{O}}$) to scene-aware descriptions that include environmental context ($\textbf{\textcolor{ctx_color}{Ctx}}$) and inferred ambient smells ($\textbf{\textcolor{inf_color}{S}}$). In the \textbf{S2I} task, expanding the semantic scope beyond basic labels improves R@5 from 22.1 to 23.0, suggesting that language context complements the missing information from the visual modality during olfactory representation training. This trend is even more evident in language-centric tasks: for \textbf{S2T}, moving from $\textbf{\textcolor{obj_color}{O}}$ to the full $\textbf{\textcolor{obj_color}{O}} + \textbf{\textcolor{ctx_color}{Ctx}} + \textbf{\textcolor{inf_color}{S}}$ template yields an improvement from 8.0 to 11.9 R@5. Notably, the inclusion of VLM-inferred smells provides the best performance, validating our hypothesis that while visual features may miss the olfactory essence of a scene, high-level language inferring can recover these unobservable cues to better align the olfactory latent space.

\paragraph{Image Bridge Ablation.}
To isolate the benefit of our three-stream architecture from the VLM-generated language guidance, we apply the Image Bridge protocol to \mname{}: we disable the ST head and route S2T/S2IT retrieval via the IS head and a frozen CLIP proxy, mirroring the NYS$^\dagger$ (adapted) baseline (Table~\ref{tab:ib_ablation}).
S2T and S2IT degrade, confirming that the gains stem from the three-stream design and not solely from VLM-enriched training data.
Notably, \mname{} with IB still outperforms NYS$^\dagger$ (adapted) with IB, showing that co-training with the ST loss also improves the IS head.
\begin{table}[t]
\begin{minipage}[t]{0.50\linewidth}
\centering
\captionsetup{width=\linewidth}
\caption{\textbf{Image Bridge (IB) ablation.} Replacing \mname{}'s native ST head with an Image Bridge proxy confirms that performance gains stem from the three-stream architecture.}
\resizebox{\linewidth}{!}{%
\renewcommand{\arraystretch}{1.09}
\begin{tabular}{l c rr rr}
\toprule
\multirow{2}{*}{\textbf{Method}} & \multirow{2}{*}{\textbf{IB}} &
\multicolumn{2}{c}{\textbf{S2T}} & \multicolumn{2}{c}{\textbf{S2IT}} \\
\cmidrule(lr){3-4}\cmidrule(lr){5-6}
& & R@5 & R@20 & R@5 & R@20 \\
\midrule
NYS$^\dagger$ (adapted) & \cmark & 8.1 & 19.2 & 18.5 & 40.7 \\
\mname{}         & \cmark & 9.1 & 18.8 & 21.4 & 43.8 \\
\mname{} (ours)         & \xmark & \textbf{11.9} & \textbf{29.2} & \textbf{23.3} & \textbf{42.5} \\
\bottomrule
\end{tabular}%
}
\label{tab:ib_ablation}

\end{minipage}
\hfill
\begin{minipage}[t]{0.47\linewidth}
\centering
\captionsetup{width=\linewidth}
\caption{\textbf{Recombination retrieval.} \textbf{Decoded Synthesis} (ours) outperforms \textbf{Raw Recombination} on zero-shot (Object, Context) pairings unseen during training.}
\resizebox{\linewidth}{!}{%
\renewcommand{\arraystretch}{1.65}
\begin{tabular}{l cccc}
\toprule
\textbf{Query Type} & R@1 & R@5 & R@10 & R@20 \\
\midrule
\makecell[l]{Raw \\ Recombination} & \textbf{3.9} & 5.9 & 7.8 & 9.8 \\ \hline
\makecell[l]{Decoded \\ Synthesis (ours)} & 2.0 & \textbf{9.8} & \textbf{11.8} & \textbf{13.7} \\
\bottomrule
\end{tabular}%
}
\label{tab:disentanglement}
\end{minipage}
\end{table}

\paragraph{VLM Sensitivity Analysis}
\label{sub:vlm_sensitivity}

\begin{table}[t]
\centering
\renewcommand{\arraystretch}{1.2}
\setlength{\tabcolsep}{4pt}
\caption{\textbf{VLM sensitivity analysis.} All variants are \mname{} trained with $\textbf{\textcolor{obj_color}{O}} + \textbf{\textcolor{ctx_color}{Ctx}} + \textbf{\textcolor{inf_color}{S}}$ descriptors; only the VLM used for annotation differs. MMLU score is reported as a proxy for general reasoning capability.}
\resizebox{0.9\textwidth}{!}{%
\begin{tabular}{l r rr rr rr}
\toprule
\multirow{2}{*}{\textbf{VLM}} & \multirow{2}{*}{\textbf{MMLU}} &
\multicolumn{2}{c}{\textbf{S2I}} & \multicolumn{2}{c}{\textbf{S2T}} & \multicolumn{2}{c}{\textbf{S2IT}} \\
\cmidrule(lr){3-4}\cmidrule(lr){5-6}\cmidrule(lr){7-8}
& & R@5 & R@20 & R@5 & R@20 & R@5 & R@20 \\
\midrule
Gemma~4~E4B~\cite{gemma2026}    & 69.4 & 19.6 & 43.3 & 7.7  & 19.8 & 20.2 & 44.4 \\
Qwen2.5-VL-72B~\cite{qwen25technicalreport} & 71.2 & 20.0 & 42.1 & 8.1  & 20.5 & 19.7 & 41.9 \\
Qwen3-VL-8B~\cite{qwen3technicalreport}    & 71.6 & 19.1 & 39.9 & 6.5  & 22.0 & 20.3 & 40.0 \\
Qwen3-VL-30B~\cite{qwen3technicalreport} (ours) & \textbf{77.8} & \textbf{23.0} & \textbf{43.6} & \textbf{11.9} & \textbf{29.2} & \textbf{23.3} & \textbf{42.5} \\
\bottomrule
\end{tabular}
}
\label{tab:vlm_sensitivity}
\end{table}

We evaluate how the choice of Vision-Language Model affects retrieval performance, keeping all other components of \mname{} fixed. Results are shown in Table~\ref{tab:vlm_sensitivity}. Performance scales with VLM reasoning quality rather than parameter count alone: Qwen3-VL-30B outperforms the larger Qwen2.5-VL-72B across all tasks, consistent with the multi-hop reasoning required to infer plausible olfactory cues from visual context. Smaller models (Qwen3-VL-8B, Gemma~4~E4B) produce weaker annotations, yielding lower retrieval performance, particularly on language-centric tasks (S2T). These results confirm that annotation richness, and therefore downstream alignment quality, is driven primarily by the model's reasoning capability rather than its scale alone.

\subsection{Annotation quality.}
\begin{figure}[t]
    \centering
    \includegraphics[width=0.85\linewidth]{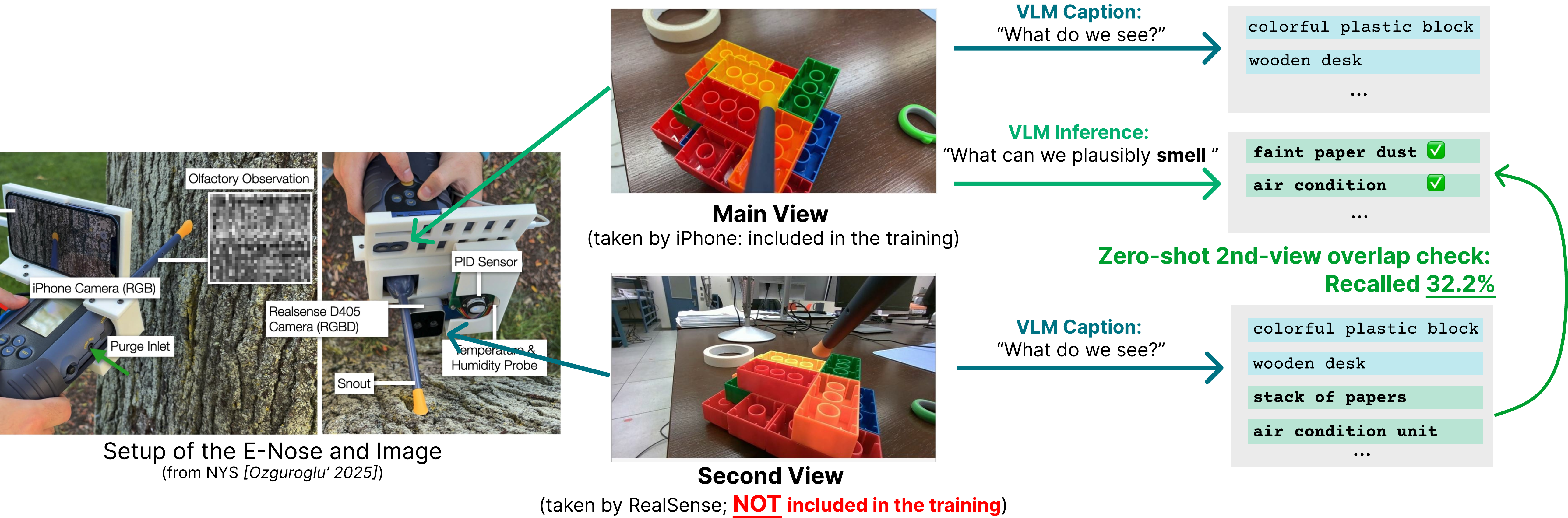}
    \caption{\textbf{Held-out view validation setup.} A VLM infers plausible smells from View~1; a second VLM judge uses a held-out View~2 to verify the inferences, distinguishing grounded predictions from hallucinations.
    }
    \label{fig:view2_validation}
\end{figure}

To verify that VLM-generated descriptors infer real olfactory context rather than plausible hallucinations, we use a second VLM, the ``judge''. We provide a held-out second view of the same scene \footnote{The NYS dataset provides two different camera views per recording, captured simultaneously from different angles of the same scene (Figure~\ref{fig:toy_ex}). View 2 is never used during the annotation process or in training. It is withheld exclusively for this validation experiment.},
and the judge scores each annotation on two criteria: \textbf{plausibility} (is the annotation contextually reasonable given both views?), reaching 98.6\%; and \textbf{View-2 confirmation} (does the annotation specifically predict content hidden in View~1 but visible in View~2?), reaching 32.2\%.
This confirms that our descriptors capture real concealed scene content rather than generic guesses, as illustrated in Figure~\ref{fig:view2_validation}.

\subsection{Latent Smell Disentanglement}

We evaluate the ability of \mname{} to decompose a complex olfactory signal into its constituent parts: the target object and the environmental context.

\paragraph{Evaluation Protocol: Recombination Retrieval.}
To evaluate whether the latent decomposition $\phi_{\text{obj}}$ and $\phi_{\text{ctx}}$ effectively isolates object-specific and environmental olfactory components, we propose a \textbf{zero-shot recombination retrieval} task. We identify (Object, Context) pairs in the validation set that never appear together in the training data, although the individual objects and contexts are present separately across different training samples.

For each such pair, we construct a synthetic query by pairing an object factor $z^S_{\text{obj}}$ from one training sample (e.g., ``coffee'') with a context factor $z^S_{\text{ctx}}$ from a distinct training sample with the target background (e.g., ``outdoor park''). We then utilize the decoder $g(\cdot)$ to synthesize a novel smell fingerprint $\hat{\mathbf{X}} = g([z^S_{\text{obj}}; z^S_{\text{ctx}}])$. The task is to retrieve the ground-truth validation samples where this specific combination occurred.

We compare our \textbf{Decoded Synthesis} against a \textbf{Raw Recombination} baseline, which creates a query by manually concatenating the raw sensor baseline ($\mathbf{B}$) and sample ($\mathbf{S}$) signals from two different training exemplars. This protocol tests the model’s ability to perform compositional generalization.

\paragraph{Quantitative Analysis.}As shown in Table~\ref{tab:disentanglement}, we evaluate the retrieval performance of these synthetic queries within our latent space, by passing the query smell fingerprint $\mathbf{X}$ through our smell encoder. Our Decoded Synthesis outperforms the Raw Recombination baseline. While the raw baseline plateaus at 9.8 R@20, our synthesized fingerprints achieve 13.7 R@20.

\section{Limitations and Conclusion}
Despite the performance gains of \mname{}, several challenges remain. The reliance on e-nose sensors introduces inherent stochasticity, as these hardware devices are prone to temporal drift and cross-sensitivity to environmental factors such as humidity and temperature. Furthermore, while the NYS dataset is a significant milestone, a substantial gap remains between olfactory benchmarks and the web-scale datasets used to train vision-language backbones. Narrowing this gap may benefit from synthetic data augmentation~\cite{boudier2025dipsy}.

In this paper, we addressed the fundamental challenge of aligning high-dimensional electronic-nose signals with visual data, recognizing that images often fail to capture the pervasive environmental context of olfaction. We introduced \mname{}, a multimodal framework that employs language guidance as a semantic bridge, leveraging the extensive world knowledge embedded in VLMs to infer plausible ambient olfactory cues. Through a language-guided latent decomposition, our model effectively disentangles object-specific odors from broader environmental factors. Extensive experiments on the New York Smells (NYS) dataset demonstrate that this semantic grounding enhances cross-modal retrieval performance across both single and joint modality tasks. Moroever, our language-guidance mechanism achieves decompositionality of olfactory information, facilitating generalization to out-of-distribution (OOD) samples.

\section*{Acknowledgements}
This work is supported by Hi! Paris and the ANR/France 2030 program (ANR-23-IACL-0005), a Hi! Paris grant and fellowship, a CIEDS grant, and a Google DeepMind academic gift. Computing resources were provided by GENCI through access to the IDRIS High-Performance Computing facilities under allocations 2026-AD011014300R3 and 2025-AD011015893R1, and by Google Gemini.
We sincerely thank Mathieu Aubry and the anonymous reviewers for their insightful discussions that contributed to this work. We are also grateful to Julie Mordacq and Robin Courant for their meticulous proofreading.

\bibliographystyle{splncs04}
\bibliography{shortstrings, main}

\newpage
\appendix
\renewcommand{\theHsection}{appendix.\thesection}
\renewcommand{\theHsubsection}{appendix.\thesubsection}
\counterwithout*{figure}{section}
\counterwithout*{table}{section}

\clearpage
\begin{center}
{\LARGE\bfseries Supplementary Material}\\[.4em]
{\large\itshape What Images Cannot Say: Language-Guided Olfactory Representation Learning}
\end{center}
\vspace{1em}

This supplementary material provides further technical depth and empirical evidence to support our main findings. Specifically, we include: (1) comprehensive implementation details; (2) details on the semantic augmentation pipeline for extracting inference smell information; (3) additional quantitative results across diverse benchmarks; (4) ablation studies on key design choices; and (5) further qualitative examples and visualizations.

\begin{enumerate}
    \renewcommand{\labelenumi}{}
    \setlength{\itemindent}{-1.5em}
    \item \ref{sec:impl_details}.~\nameref{sec:impl_details}
    \item \ref{sec:vlm_augmentation}.~\nameref{sec:vlm_augmentation}
    \begin{enumerate}
        \renewcommand{\labelenumii}{}
        \setlength{\itemindent}{-1.5em}
        \item \ref{subsec:annotations}.~\nameref{subsec:annotations}
        \item \ref{sub:annotation_stats}.~\nameref{sub:annotation_stats}
    \end{enumerate}
    \item \ref{sec:Quant_Results}.~\nameref{sec:Quant_Results}
    \begin{enumerate}
        \renewcommand{\labelenumii}{}
        \setlength{\itemindent}{-1.5em}
        \item \ref{sub:retrieval}.~\nameref{sub:retrieval}
        \item \ref{sub:classification}.~\nameref{sub:classification}
        \item \ref{sub:depth}.~\nameref{sub:depth}
    \end{enumerate}
    \item \ref{sec:ablations}.~\nameref{sec:ablations}
    \begin{enumerate}
        \renewcommand{\labelenumii}{}
        \setlength{\itemindent}{-1.5em}
        \item \ref{sub:llm_ablation}.~\nameref{sub:llm_ablation}
        \item \ref{sub:fusion_strategies}.~\nameref{sub:fusion_strategies}
        \item \ref{sub:twostage}.~\nameref{sub:twostage}
        \item \ref{sub:high_granularity}.~\nameref{sub:high_granularity}
    \end{enumerate}
    \item \ref{sec:Quali_Results}.~\nameref{sec:Quali_Results}
\end{enumerate}

\section{Implementation Details}
\label{sec:impl_details}

\paragraph{Dataset and Custom Splits.}
Since the original splits for the New York Smells (NYS) dataset~\cite{ozguroglu2025new} have not been released, we established an independent partitioning to facilitate our experiments. We use a training set of 5,996 samples and a validation set of 936 samples. All reported results, including the reproduced NYS baseline and our proposed \mname{}, are evaluated using these identical splits to ensure a fair and consistent comparison.

\paragraph{Baseline Reproduction.}
To ensure a rigorous evaluation, we reproduce the NYS baseline \cite{ozguroglu2025new} using our standardized data splits and the same Transformer-based smell encoder architecture employed in \mname{}. By adjusting the backbone design and optimization strategy, our reproduced baseline achieves performance metrics that exceed those originally reported in \cite{ozguroglu2025new}, providing a more challenging and fair point of comparison for our multimodal approach.

\paragraph{Architectural Specifications.}
Our smell encoder $f_S$ is a Transformer with a model dimension $d=448$, 6 layers, and 8 attention heads. The input e-nose signal consists of 32 sensor channels across 28 timesteps (14 for the baseline $\mathbf{B}$ and 14 for the sample $\mathbf{S}$). While most recordings in the NYS dataset follow this duration, any samples exceeding 14 seconds per phase are truncated to maintain a uniform temporal resolution. This input is projected into the embedding space via a linear layer and augmented with sinusoidal positional encodings. For the visual and textual backbones, we utilize CLIP ViT-B/16 \cite{radford2021learning}. While the textual encoder remains frozen to preserve its pre-trained semantic structure, the vision encoder is fine-tuned to align its representations with the olfactory features. The CLIP vision backbone contains 85.8M trainable parameters, while the smell encoder contains 14.9M trainable parameters. The two modality-specific projection heads, $\phi_I$ and $\phi_T$, are implemented as MLPs that map the $448$-dimensional olfactory vector into the $512$-dimensional CLIP latent space.

\paragraph{Training and Optimization.}
Models are trained using the AdamW optimizer with a batch size of 512. To improve the robustness of the visual representations, we apply strong random resized crops and horizontal flips during training. All training is conducted in FP16 mixed precision on a single NVIDIA H100 GPU.

\section{Semantic Augmentation Details}
\label{sec:vlm_augmentation}

\paragraph{Terminology Clarification.} To avoid confusion with the original NYS taxonomy \cite{ozguroglu2025new}, we distinguish between the \textit{closed-set} categories used for evaluation and the \textit{open-set} descriptors used for model training. 
\begin{itemize}
    \item \textbf{\textcolor{obj_color}{Object (O)}, \textcolor{ctx_color}{Context (Ctx)} \& \textbf{\textcolor{inf_color}{Smell (S)}}:} Open-set natural language descriptors generated by our VLM pipeline. These are used directly as linguistic supervision in \mname{}.
    \item \textbf{Item (I) \& Environment (E):} Closed-set categorical labels (43 and 7 classes, respectively) following the taxonomy shown in \cite{ozguroglu2025new}. These are used \textbf{only} for the classification probing tasks in Section~\ref{sub:classification} of the supplementary material, and nowhere else.
\end{itemize}

\subsection{Prompt Templates}
\label{subsec:annotations}
We utilize Qwen3VL-30B \cite{qwen3technicalreport} to extract semantic anchors from scene images. To ensure the generated descriptors are relevant, we use a structured prompt that forces the model to decouple the primary olfactory information.

\paragraph{Usage of Descriptors.}
While the VLM generates five distinct fields, they serve different roles in our study. The \textit{open-set} \textbf{\textcolor{obj_color}{Object (O)}} and \textbf{\textcolor{ctx_color}{Context (Ctx)}}, as well as the inferred \textbf{\textcolor{inf_color}{Smell (S)}} descriptors are used to construct the language-guided supervision for \mname{}. The \textit{closed-set} \textbf{Item} and \textbf{Environment} fields are used exclusively as ground-truth labels for the linear probing tasks (Section~\ref{sub:classification}) to maintain compatibility with the original NYS taxonomy.

\paragraph{Primary VLM Prompt.}
\begin{quote}
\ttfamily \small \raggedright
Analyze this image for environmental olfactory context.  The yellow tip of the probe (the "snout") is sampling a "smell print" from a specific object or surface.

\vspace{.5cm}

1. ITEM (I): The general category of the object the yellow tip is touching. Choose ONE: \{items\_list\}.

2. OBJECT (O): A specific, open-set name for the exact object or surface the yellow tip is touching.

3. ENVIRONMENT (E): The general environment/setting. Choose ONE: \{environments\_list\}.

4. CONTEXT (Ctx): A short, open-set phrase describing the background/setting (e.g. park, garage, kitchen, office, street, garden, workshop). Anything that describes where the scene is.

5. INFERRED SMELLS (S): Infer likely smells from the surrounding environment ONLY. 

***CRITICAL: EXCLUDE the smell of the OBJECT (O) itself.

*** Focus on "invisible" scents that are likely in the air (e.g., traffic exhaust, humidity, air conditioning, distant greenery). 

\vspace{.5cm}

Respond in this exact format (one field per line):

ITEM: [exactly one from the item list]

OBJECT: [name]

ENVIRONMENT: [exactly one from the environment list]

CONTEXT: [one short open-set phrase]

SMELLS: [ambient smell 1], [ambient smell 2]

\vspace{.5cm}

---

EXAMPLE:

ITEM: plants flower ornamental

OBJECT: white daffodil

ENVIRONMENT: Campus Outdoors

BACKGROUND: a breezy concrete walkway adjacent to a freshly mowed lawn

SMELLS: fresh cut grass, concrete dust, distant vehicle exhaust

---
\end{quote}
\paragraph{Scene Inference with Different Language Models (Two-Stage Pipeline).}
In Section~\ref{sub:llm_ablation}, we evaluate an alternative pipeline to determine whether olfactory reasoning is best performed by grounded vision-language models or via decoupled language reasoning. In this variant, we replace the single-stage VLM inference with a two-stage process:

\begin{enumerate}
    \item \textbf{Stage 1 (Visual Captioning)}: The VLM generates a dense, generic description of the scene image.
    \item \textbf{Stage 2 (Linguistic Reasoning)}: This description is passed to a separate Large Language Model (\textbf{Qwen3-30B} \cite{qwen3technicalreport}) which performs the final olfactory inference based on the text alone.
\end{enumerate}

By decoupling visual perception from olfactory inference, we can test whether olfactory knowledge is more effectively extracted via direct visual grounding or through the high-level semantic priors encoded in LLMs. The prompts used for this two-stage pipeline are provided below:

\paragraph{Stage 1 VLM Prompt:}
\begin{quote}
\ttfamily \small \raggedright
Caption this image. Ignore the blue and yellow tool.
\end{quote}

\paragraph{Stage 2 LLM Prompt:}
\begin{quote}
\ttfamily \small \raggedright
You are given a short textual description of an image. Based only on this description, provide four outputs:

\vspace{.5cm}

1. ITEM (I): The main item/object that would be shown. Choose exactly ONE from this list: \{items\_list\}

2. OBJECT (O): A specific, open-set name for the exact object.

3. ENVIRONMENT (E): The general environment/setting. Choose exactly ONE from this list: \{environments\_list\}

4. CONTEXT (Ctx): A short, open-set phrase describing the background/setting (e.g. park, garage, kitchen, office, street, garden, workshop). Anything that describes where the scene is.

5. INFERRED SMELLS (S): Infer likely smells from the surrounding environment ONLY. 

***CRITICAL: EXCLUDE the smell of the OBJECT (O) itself.

*** Focus on 'invisible' scents that are likely in the air (e.g., traffic exhaust, humidity, air conditioning, distant greenery). 

\vspace{.5cm}

Respond in this exact format (one field per line):

ITEM: [exactly one from the item list]

OBJECT: [name]

ENVIRONMENT: [exactly one from the environment list]

CONTEXT: [one short open-set phrase]

SMELLS: [ambient smell 1], [ambient smell 2]

\vspace{.5cm}

---

EXAMPLE:

ITEM: plants flower ornamental

OBJECT: white daffodil

ENVIRONMENT: Campus Outdoors

CONTEXT: a breezy concrete walkway adjacent to a freshly mowed lawn

SMELLS: fresh cut grass, concrete dust, distant vehicle exhaust

---

\vspace{.5cm}

Description of the image: \{caption\}
\end{quote}

\subsection{Annotation Statistics}
\label{sub:annotation_stats}

\begin{figure}[htp!]
    \centering
    \begin{subfigure}[t]{0.5\textwidth}
        \centering
        \includegraphics[width=\textwidth]{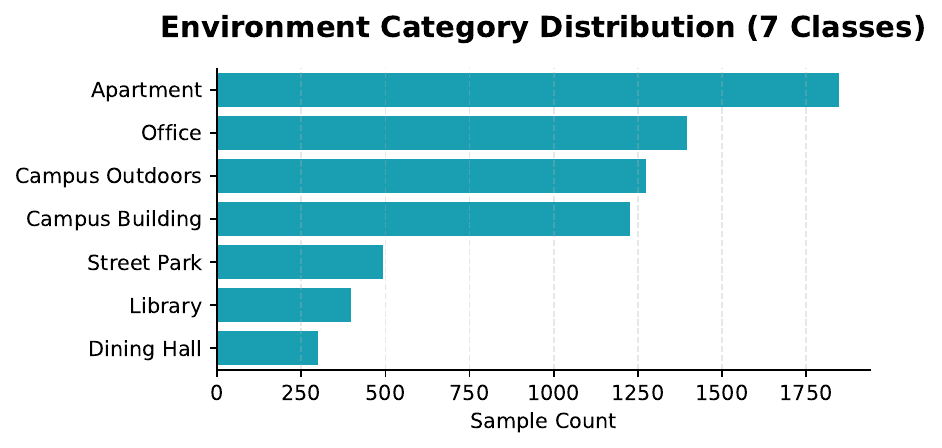}
        \caption{Environment Categories (7 classes)}
    \end{subfigure}

    \vspace{0.6em}

    \begin{subfigure}[t]{0.5\textwidth}
        \centering
        \includegraphics[width=\textwidth]{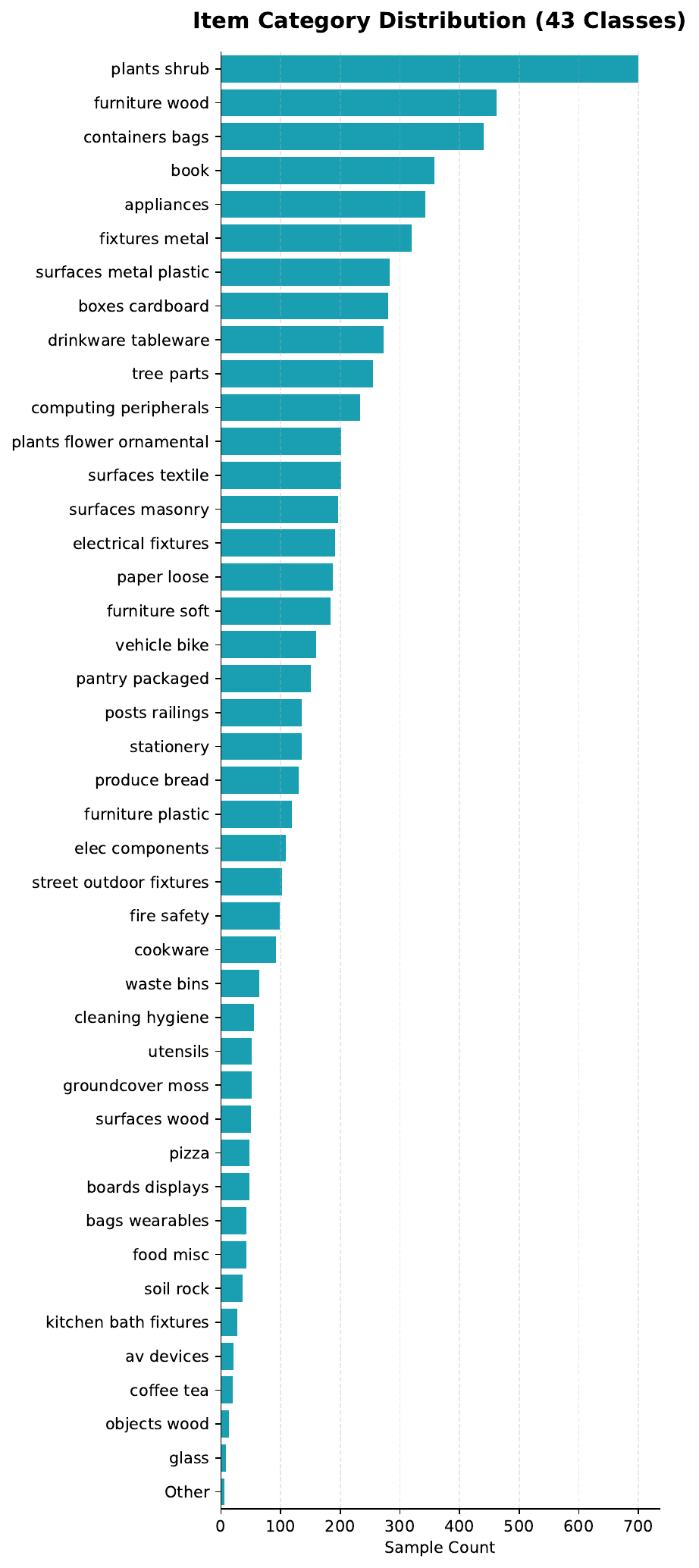}
        \caption{Item Categories (43 classes)}
    \end{subfigure}
    
    \caption{\textbf{Taxonomy distribution of the NYS benchmark.} We visualize the label frequency for the two classification probing tasks: (a) the 7 Environment categories and (b) the 43 Item categories. Note the long-tail distribution in the Item classes, which challenges the model to learn robust categorical representations from imbalanced e-nose signals.}
    \label{fig:dataset_stats}
\end{figure}

We analyze the distribution of the VLM-generated pseudo-labels used for our probing tasks. As shown in Figure~\ref{fig:dataset_stats}, the taxonomy covers a broad spectrum of in-the-wild olfactory scenes.

\section{Quantitative results}
\label{sec:Quant_Results}
\subsection{Inverse Retrieval Tasks: I2S and T2S}
\label{sub:retrieval}

To further validate the bidirectional consistency of our learned latent spaces, we evaluate \mname{} on the inverse retrieval tasks: \textbf{Image-to-Smell (I2S)} and \textbf{Text-to-Smell (T2S)}. These experiments assess the model's ability to identify a specific smell fingerprint given a visual or textual query.

\paragraph{Baselines and Protocols.}
As with the forward tasks, we compare against a reproduced version of the NYS baseline \cite{ozguroglu2025new}. For the \textbf{T2S} task, since the baseline lacks a native textual head, we implement the \textbf{Image Bridge} protocol described in Section~4.1 of the main paper: a query text is first mapped to its most similar image in the gallery via zero-shot CLIP \cite{radford2021learning} similarity, and this retrieved image is then used to query the olfactory database.

\paragraph{Analysis of Results.} As shown in Table~\ref{tab:exp3_results}, \mname{} outperforms the baseline in \textbf{I2S} retrieval, improving R@5 from 18.5 to 22.3. This confirms that the semantic information provided by our VLM-augmented language guidance (O+Ctx+S) during training helps the olfactory encoder learn features that are more effectively aligned with the visual manifold.

In the \textbf{T2S} task, we observe that the \textbf{NYS (adapted)} baseline achieves slightly higher recall (9.9 R@5) than our direct T2S approach (7.2 R@5). This result is expected due to the structural nature of the Image Bridge protocol. Specifically, the adapted baseline leverages the massive, web-scale prior of the frozen CLIP model to align text queries with visual candidates before performing the final I2S step. In contrast, our model performs \textbf{direct native retrieval} without ever accessing the image modality during the query process.

The fact that our model maintains competitive performance ($22.6\%$ R@20) while relying solely on the alignment between the e-nose signal and textual descriptors demonstrates the strength of our native olfactory-textual projection. While a visual bridge provides a shortcut for retrieval, our end-to-end alignment proves that the olfactory encoder is capable of capturing high-level semantic concepts purely through linguistic supervision.

\begin{table}[t]
\centering
\caption{\textbf{Inverse retrieval performance on the NYS dataset.} We evaluate the bidirectional consistency of our olfactory latent space via Image-to-Smell (I2S) and Text-to-Smell (T2S) tasks. For T2S, the \textbf{NYS (adapted)} baseline utilizes an \textit{Image Bridge} protocol, leveraging CLIP's pre-existing image-text alignment to rank candidates. In contrast, \textbf{\mname{} (Ours)} performs direct retrieval using its native textual projection head. Our method outperforms the baseline in I2S and maintains competitive performance in T2S without requiring visual proxies during the query process.}
\renewcommand{\arraystretch}{1.2}
\setlength{\tabcolsep}{3pt}
\resizebox{\textwidth}{!}{%
\setlength{\tabcolsep}{4pt}
\begin{tabular}{l ccc *{4}{r} *{4}{r}}
\toprule
\multirow{2}{*}{\textbf{Method}} & \multicolumn{3}{c}{\textbf{Text Used}} & \multicolumn{4}{c}{\textbf{I2S}} & \multicolumn{4}{c}{\textbf{T2S}} \\
\cmidrule(lr){2-4} \cmidrule(lr){5-8} \cmidrule(lr){9-12}
& $\textbf{\textcolor{obj_color}{O}}$ & $\textbf{\textcolor{ctx_color}{Ctx}}$ & $\textbf{\textcolor{inf_color}{S}}$ & R@1 & R@5 & R@10 & R@20 & R@1 & R@5 & R@10 & R@20 \\
\midrule

NYS (reproduced) & & & & 4.2 & 18.5 & 28.2 & 39.6 & \multicolumn{4}{c}{---} \\
\midrule

NYS (adapted) & \checkmark & \checkmark & \checkmark & \multicolumn{4}{c}{---} & \tightbold{2.1} & \tightbold{9.9} & \tightbold{16.7} & \tightbold{26.0} \\
\midrule

Ours & \checkmark & \checkmark & \checkmark & \tightbold{5.3} & \tightbold{22.3} & \tightbold{32.9} & \tightbold{46.2} & 1.2 & 7.2 & 12.8 & 22.6 \\
\bottomrule
\end{tabular}%
}
\label{tab:exp3_results}
\end{table}

\subsection{Classification}
\label{sub:classification}

To further evaluate the richness of the learned olfactory representations, we perform a linear probing analysis on two classification tasks: \textbf{Item} and \textbf{Environment}. Since the original NYS dataset \cite{ozguroglu2025new} provides raw images and sensor data without discrete categorical labels, we utilize a pretrained VLM (Qwen3VL-30B \cite{qwen3technicalreport}) to generate a set of closed-set pseudo-labels for the entire dataset.

\paragraph{Label Generation.} We prompt the VLM to classify each scene into one of the 43 item categories and 7 environment types defined in the taxonomy of the NYS benchmark. The specific prompts used for this labeling process are detailed in Section~\ref{subsec:annotations}. These VLM-generated labels serve as the ground truth for our classification experiments.

\paragraph{Linear Probing Protocol.} We freeze the parameters of the smell encoder $f_S$ to ensure that no further representational learning occurs during this task. We extract the penultimate representation $z^S$ (the encoder output prior to the modality-specific projection heads) for all samples in the training and validation sets. A single linear layer is then trained as a probe for each task using a standard cross-entropy loss and the Adam optimizer. This protocol measures the degree to which item-specific and context-specific semantic information is linearly separable within the fixed olfactory latent space.

\paragraph{Results.} We report Top-1 Accuracy on the held-out validation set. As shown in Table~\ref{tab:item_env_results}, we evaluate the categorical richness of the learned olfactory latent space through linear and non-linear (MLP) probing. Both \mname{} and the NYS baseline significantly outperform the chance-level baseline, confirming that e-nose signals carry substantial discriminative information about both the target object and its surroundings.

When employing a linear probe, \mname{} reaches 18.48\% and 48.08\% accuracy for Item and Environment classification, respectively. The introduction of a non-linear MLP probe yields a performance boost across both tasks, with our method achieving scores of 25.43\% (Item) and 56.84\% (Environment). This suggests that our multimodal alignment, incorporating VLM-augmented language descriptors, produces a feature space that is rich in semantic content and resilient to the inherent noise of urban olfactory measurements.

\begin{table}[t]
\centering
\renewcommand{\arraystretch}{1.3}
\setlength{\tabcolsep}{6pt}
\caption{\textbf{Probing olfactory representations.} We evaluate the linear and non-linear (MLP) separability of the frozen smell encoder $f_S$ across two classification tasks: \textbf{Item Category} (43 classes) and \textbf{Environmental Context} (7 classes). Both models utilize VLM-generated closed-set pseudo-labels as ground truth. While both methods significantly outperform the chance baseline, \mname{} achieves superior performance in \textbf{Environmental} classification (56.84\%), demonstrating that our multimodal alignment creates a latent space that better captures the global semantic context of the olfactory scene.}
\resizebox{.6\textwidth}{!}{
\begin{tabular}{l ccc}
\toprule
\textbf{Method} & \textbf{Probe} & \multicolumn{1}{c}{\textbf{Item}} & \multicolumn{1}{c}{\textbf{Environment}} \\
\cmidrule(lr){3-3} \cmidrule(lr){4-4}
& & {Acc} & {Acc} \\
\midrule
Chance & --- & 2.33 & 14.29 \\
\midrule
\multirow{2}{*}{NYS} & Linear & 18.16 & 50.11 \\
& MLP & \textbf{26.71} & 55.88 \\
\midrule
\multirow{2}{*}{Ours} & Linear & 18.48 & 48.08 \\
& MLP & 25.43 & \textbf{56.84} \\
\bottomrule
\end{tabular}
}
\label{tab:item_env_results}
\end{table}

\subsection{Performance by Scene Depth}
\label{sub:depth}

Although \mname{} does not use depth as an input, the NYS dataset provides co-registered depth measurements for every sample. We exploit this to probe whether retrieval performance is affected by the physical geometry of the captured scene. Specifically, we partition the validation set into \textbf{enclosed} and \textbf{open} environments based on the depth reading, and evaluate each subset against its own gallery.

As shown in Table~\ref{tab:depth}, \mname{} outperforms the NYS$^\dagger$ baseline in both subsets across all tasks and metrics. The improvement is larger for open scenes, where \mname{} achieves the best performance overall. We attribute this to richer visual context in open environments: a VLM observing a street scene or a park can infer a more diverse and precise set of ambient smells, producing stronger language supervision and better downstream smell--scene alignment. Enclosed scenes, by contrast, offer less contextual diversity and may constrain the range of inferred ambient odors. Designing annotation strategies that remain effective in visually sparse, enclosed environments is a promising direction for future work.

\begin{table}[t]
\centering\small
\setlength{\tabcolsep}{4pt}
\caption{\textbf{Performance by scene depth (enclosed vs.\ open).} We split the validation set by depth into enclosed and open scenes and evaluate each subset against its own gallery. \mname{} improves over the baseline in both subsets, with the largest gains on open scenes.}
\resizebox{0.75\columnwidth}{!}{%
\begin{tabular}{llcccccc}
\toprule
& & \multicolumn{2}{c}{S2I} & \multicolumn{2}{c}{S2T} & \multicolumn{2}{c}{S2IT} \\
\cmidrule(lr){3-4}\cmidrule(lr){5-6}\cmidrule(lr){7-8}
Method & Subset & R@5 & R@20 & R@5 & R@20 & R@5 & R@20 \\
\midrule
\multirow{3}{*}{NYS$^\dagger$} & all      & 20.0 & 42.0 & 8.1  & 19.2 & 20.0 & 40.9 \\
                               & enclosed & 19.4 & 41.2 & 6.9  & 19.0 & 19.4 & 40.9 \\
                               & open     & 20.0 & 42.5 & 9.2  & 19.1 & 20.0 & 40.6 \\
\midrule
\multirow{3}{*}{\mname{} (ours)} & all      & 23.0 & 43.6 & 11.9 & \textbf{29.2} & 23.3 & 42.5 \\
                               & enclosed & 20.7 & 42.7 & 10.8 & 29.0          & 21.4 & 41.8 \\
                               & open     & \textbf{24.9} & \textbf{44.2} & \textbf{13.0} & 29.1 & \textbf{24.7} & \textbf{42.9} \\
\bottomrule
\end{tabular}
}
\label{tab:depth}
\end{table}

\section{Ablation Studies}
\label{sec:ablations}

\subsection{Scene Inference with Different Language Models}
\label{sub:llm_ablation}

\begin{table}[t]
\centering
\renewcommand{\arraystretch}{1.2}
\caption{\textbf{Ablation on Scene Inference with Different Language Models: Visual Grounding vs. Two-Stage LLM Pipeline.} We compare the performance of \mname{} when trained using semantics directly from a VLM (Qwen3-VL) versus those generated by an LLM (Qwen3-30B) based on visual captions. This comparison measures the trade-off between the direct visual grounding of a VLM and the common-sense priors of a large-scale instruction-tuned LLM.}
\setlength{\tabcolsep}{6pt}
\resizebox{\textwidth}{!}{%
\begin{small}
\begin{tabular}{l ccc cc cc cc}
\toprule
\multirow{2}{*}{\textbf{Method}} & \multicolumn{3}{c}{\textbf{Textual Info}} & \multicolumn{2}{c}{\textbf{S2I}} & \multicolumn{2}{c}{\textbf{S2T}} & \multicolumn{2}{c}{\textbf{S2IT}} \\
\cmidrule(lr){2-4} \cmidrule(lr){5-6} \cmidrule(lr){7-8} \cmidrule(lr){9-10}
& $\textbf{\textcolor{obj_color}{O}}$ & $\textbf{\textcolor{ctx_color}{Ctx}}$ & $\textbf{\textcolor{inf_color}{S}}$ & R@5 & R@20 & R@5 & R@20 & R@5 & R@20 \\
\midrule

\multirow{3}{*}{Ours (VLM)} 
& \checkmark &            &            & 22.1 & 42.4 & 8.0 & 21.5 & 22.0 & 42.6 \\
& \checkmark & \checkmark &            & 21.8 & \textbf{45.2} & 10.7 & 28.1 & 21.3 & \textbf{45.7} \\
& \checkmark & \checkmark & \checkmark & \textbf{23.0} & 43.6 & \textbf{11.9} & \textbf{29.2} & \textbf{23.3} & 42.5 \\
\midrule

\multirow{3}{*}{Ours (LLM)} 
& \checkmark &            &            & 21.2 & 42.4 & 6.6 & 18.2 & 21.5 & 42.5 \\
& \checkmark & \checkmark &            & 21.9 & 42.7 & 8.7 & 24.1 & 22.1 & 42.9 \\
& \checkmark & \checkmark & \checkmark & 22.0 & 44.1 & 9.1 & 23.7 & 22.2 & 44.2 \\
\bottomrule
\end{tabular}
\end{small}
}
\label{tab:llm_ablation_results}
\end{table}

We investigate the impact of the semantic source on olfactory representation learning by comparing our primary VLM-driven pipeline (Ours (VLM)) with a two-stage decoupled pipeline (Ours (LLM)) utilizing Qwen3-30B for reasoning. As shown in Table~\ref{tab:llm_ablation_results}, we observe two key trends:

\paragraph{The Superiority of Visual Grounding.} Across all retrieval tasks, \mname{} with VLM-generated annotations consistently outperforms the LLM-based variant. For the most complete language setting ($\textbf{\textcolor{obj_color}{O}} + \textbf{\textcolor{ctx_color}{Ctx}} + \textbf{\textcolor{inf_color}{S}}$), the VLM pipeline achieves 29.2\% R@20 in S2T retrieval, compared to 23.7\% for the LLM pipeline. This gap suggests that when a VLM directly "sees" the scene, it captures nuanced environmental cues that are lost when the scene is first compressed into a generic text caption for an LLM. Visual grounding ensures that the inferred ambient smells ($\textbf{\textcolor{inf_color}{S}}$) are spatially and contextually anchored to the specific instance, rather than being generalized "common sense" guesses.

\paragraph{Robustness of the Two-Stage Pipeline.}
Despite the performance gap, the LLM pipeline remains highly competitive, particularly in the S2I task (44.1\% R@20). The fact that the LLM-based pipeline still outperforms the vision-only NYS baseline (42.0\% R@20) demonstrates that the "olfactory common sense" encoded in large language models is a powerful prior. Even without direct visual access, the LLM is able to successfully infer a plausible olfactory manifold based solely on a textual description of the scene. However, for precise multimodal alignment, the end-to-end visual grounding provided by the VLM remains the optimal choice.

\subsection{Fusion Strategies for Joint-Modality Retrieval (S2IT)}
\label{sub:fusion_strategies}
To perform joint retrieval, we evaluate two distinct fusion paradigms to unify the image-aligned ($z^S_I$) and textual-aligned ($z^S_T$) smell embeddings.
\begin{enumerate}
    \item \textbf{Latent-level Fusion}: We construct a single blended representation $\tilde{z}$ before computing similarity. This is our primary method:$$ \tilde{z}_i = \frac{\alpha\, z^{S}_{I,i} + (1-\alpha)\, z^{S}_{T,i}}{\left\lVert \alpha\, z^{S}_{I,i} + (1-\alpha)\, z^{S}_{T,i} \right\rVert_2} $$The retrieval score is then calculated by comparing this unified vector to both targets:$$ \text{score}_{ij} = \alpha \cdot \text{sim}(\tilde{z}_i, z^I_j) + (1-\alpha) \cdot \text{sim}(\tilde{z}_i, z^T_j) $$
    \item \textbf{Similarity-level Fusion}: A late-fusion approach where the joint score is a weighted combination of independent similarity scores:$$ \text{score}_{ij} = \alpha \cdot \text{sim}(z^S_{I,i}, z^I_j) + (1-\alpha) \cdot \text{sim}(z^S_{T,i}, z^T_j) $$
\end{enumerate}
Latent-level fusion forces the model to find a single point in the latent manifold that satisfies both visual and linguistic constraints simultaneously. Unless otherwise specified, we use Latent-level fusion for all joint retrieval experiments, with the fusion weight $\alpha$ selected via grid search on the validation set.

\begin{table}[t]
\centering
\caption{\textbf{Ablation of fusion strategies and modality weighting.} We evaluate the sensitivity of joint-modality retrieval (S2IT) to different fusion paradigms and visual-textual weighting coefficients ($a$). Results are reported using the full semantic template ($\textbf{\textcolor{obj_color}{O}}+\textbf{\textcolor{ctx_color}{Ctx}}+\textbf{\textcolor{inf_color}{S}}$).}
\vspace{-0.2cm}
\setlength{\tabcolsep}{10pt}
\begin{tabular}{l c cccc}
\toprule
\textbf{Fusion} & \textbf{$a$} & R@1 & R@5 & R@10 & R@20 \\
\midrule
\multirow{3}{*}{Similarity-level} 
 & 0.3 & 5.0 & 20.3 & 32.5 & 44.3 \\
 & 0.5 & 5.8 & \textbf{22.1} & 32.2 & 44.1 \\
 & 0.7 & \textbf{6.3} & 22.0 & \textbf{34.0} & \textbf{44.6} \\
\midrule
\multirow{3}{*}{Latent-level} 
 & 0.3 & 4.7 & 17.5 & 27.6 & 40.7 \\
 & 0.5 & 6.1 & 20.8 & 31.2 & 42.9 \\
 & 0.7 & \textbf{6.2} & \textbf{22.9} & \textbf{32.9} & \textbf{42.9} \\
\bottomrule
\end{tabular}
\label{tab:fusion_ablation}
\end{table}
\paragraph{Ablation.} We evaluate the sensitivity of our joint retrieval (S2IT) to different fusion paradigms and modality weightings ($a$). As shown in Table~\ref{tab:fusion_ablation}, both Similarity-level and Latent-level fusion exhibit a similar upward trend as the visual weighting $a$ increases. While Similarity-level fusion provides slightly better performance at lower visual weights, Latent-level fusion demonstrates a more significant performance gain as the modalities are balanced, reaching an R@5 of 22.9. This suggests that while Similarity fusion acts as a "greedy" ensemble of independent heads, our chosen Latent-level fusion succeeds in forcing the olfactory embedding into a singular, semantically coherent point in the CLIP manifold that satisfies both visual and textual constraints simultaneously.

\subsection{Two-Stage Training Strategy}
\label{sub:twostage}

We investigate whether warming up smell-image alignment before introducing textual supervision benefits training. In our two-stage schedule, Phase~1 trains with only the S2I loss active. At the midpoint, the S2T loss is switched on and training continues to completion with both losses active.

The two-stage schedule falls short of joint training across all tasks and metrics, as shown in Table~\ref{tab:twostage_ablation}. The S2T gap is the most pronounced ($-2.2$ R@5), but S2I and S2IT also decline. This outcome suggests that the olfactory encoder benefits from simultaneous co-alignment pressure from both objectives from the very start of training. When Phase~1 optimizes exclusively for the visual manifold, the shared encoder backbone develops representations biased toward visual features. Introducing the textual loss mid-training then forces the textual projection head to adapt to an already-committed encoder, limiting its ability to discover a complementary textual alignment. Joint training, by contrast, allows the encoder to continuously balance both objectives, yielding stronger olfactory representations.

\begin{table}[t]
\centering
\caption{\textbf{Ablation of two-stage training strategy.} We compare standard joint training against a two-stage schedule that disables the S2T loss for the first half of training before switching it on at full weight. Results use the full semantic template ($\textbf{\textcolor{obj_color}{O}}+\textbf{\textcolor{ctx_color}{Ctx}}+\textbf{\textcolor{inf_color}{S}}$).}
\vspace{-0.2cm}
\setlength{\tabcolsep}{4pt}
\resizebox{\textwidth}{!}{%
\renewcommand{\arraystretch}{1.09}
\begin{tabular}{l c rrrr rrrr rrrr}
\toprule
\multirow{2}{*}{\textbf{Method}} & \multirow{2}{*}{\textbf{\makecell{Two-\\Stage}}} & \multicolumn{4}{c}{\textbf{S2I}} & \multicolumn{4}{c}{\textbf{S2T}} & \multicolumn{4}{c}{\textbf{S2IT}} \\
\cmidrule(lr){3-6} \cmidrule(lr){7-10} \cmidrule(lr){11-14}
& & R@1 & R@5 & R@10 & R@20 & R@1 & R@5 & R@10 & R@20 & R@1 & R@5 & R@10 & R@20 \\
\midrule
\mname{}         & \cmark & 5.0 & 19.4 & 30.2 & 41.2 & 2.4 & 9.7 & 16.6 & 25.1 & 5.6 & 18.9 & 30.4 & 41.5 \\
\mname{} (ours)  & \xmark & \tightbold{6.0} & \tightbold{23.0} & \tightbold{33.5} & \tightbold{43.6} & \tightbold{3.3} & \tightbold{11.9} & \tightbold{19.8} & \tightbold{29.2} & \tightbold{6.8} & \tightbold{23.3} & \tightbold{32.7} & \tightbold{42.5} \\
\bottomrule
\end{tabular}%
}
\label{tab:twostage_ablation}
\end{table}

\subsection{Annotation Granularity}
\label{sub:high_granularity}

We test whether eliciting more inferred smells per image improves retrieval. Using the same VLM, we re-annotate the dataset with a prompt that yields a higher number of ambient smell descriptors per image rather than our default prompt. As shown in Table~\ref{tab:high_granularity}, performance degrades consistently across all tasks. We attribute this to annotation quality collapsing past the VLM's natural saturation point: once the most salient, visually-grounded smells have been listed, additional entries are increasingly speculative. Training against these low-confidence descriptors introduces label noise.

\begin{table}[t]
\centering
\caption{\textbf{Ablation on smell annotation granularity.} We compare our default prompt against a high-granularity variant that elicits a larger number of inferred ambient smells per image using the same VLM. Both variants use the full $\textbf{\textcolor{obj_color}{O}}+\textbf{\textcolor{ctx_color}{Ctx}}+\textbf{\textcolor{inf_color}{S}}$ template and identical architecture.}
\vspace{-0.2cm}
\setlength{\tabcolsep}{4pt}
\resizebox{\textwidth}{!}{%
\renewcommand{\arraystretch}{1.09}
\begin{tabular}{l c rrrr rrrr rrrr}
\toprule
\multirow{2}{*}{\textbf{Method}} & \multirow{2}{*}{\textbf{\makecell{Increased\\Inferred Smells}}} & \multicolumn{4}{c}{\textbf{S2I}} & \multicolumn{4}{c}{\textbf{S2T}} & \multicolumn{4}{c}{\textbf{S2IT}} \\
\cmidrule(lr){3-6} \cmidrule(lr){7-10} \cmidrule(lr){11-14}
& & R@1 & R@5 & R@10 & R@20 & R@1 & R@5 & R@10 & R@20 & R@1 & R@5 & R@10 & R@20 \\
\midrule
\mname{}         & \cmark & 4.9 & 19.7 & 28.6 & 41.1 & 2.4 & 7.1 & 12.4 & 20.1 & 4.9 & 19.9 & 29.2 & 40.3 \\
\mname{} (ours)  & \xmark & \tightbold{6.0} & \tightbold{23.0} & \tightbold{33.5} & \tightbold{43.6} & \tightbold{3.3} & \tightbold{11.9} & \tightbold{19.8} & \tightbold{29.2} & \tightbold{6.8} & \tightbold{23.3} & \tightbold{32.7} & \tightbold{42.5} \\
\bottomrule
\end{tabular}%
}
\label{tab:high_granularity}
\end{table}

\section{Qualitative Results}
\label{sec:Quali_Results}

\begin{figure}[htp!]
\centering
\includegraphics[width=.75\linewidth]{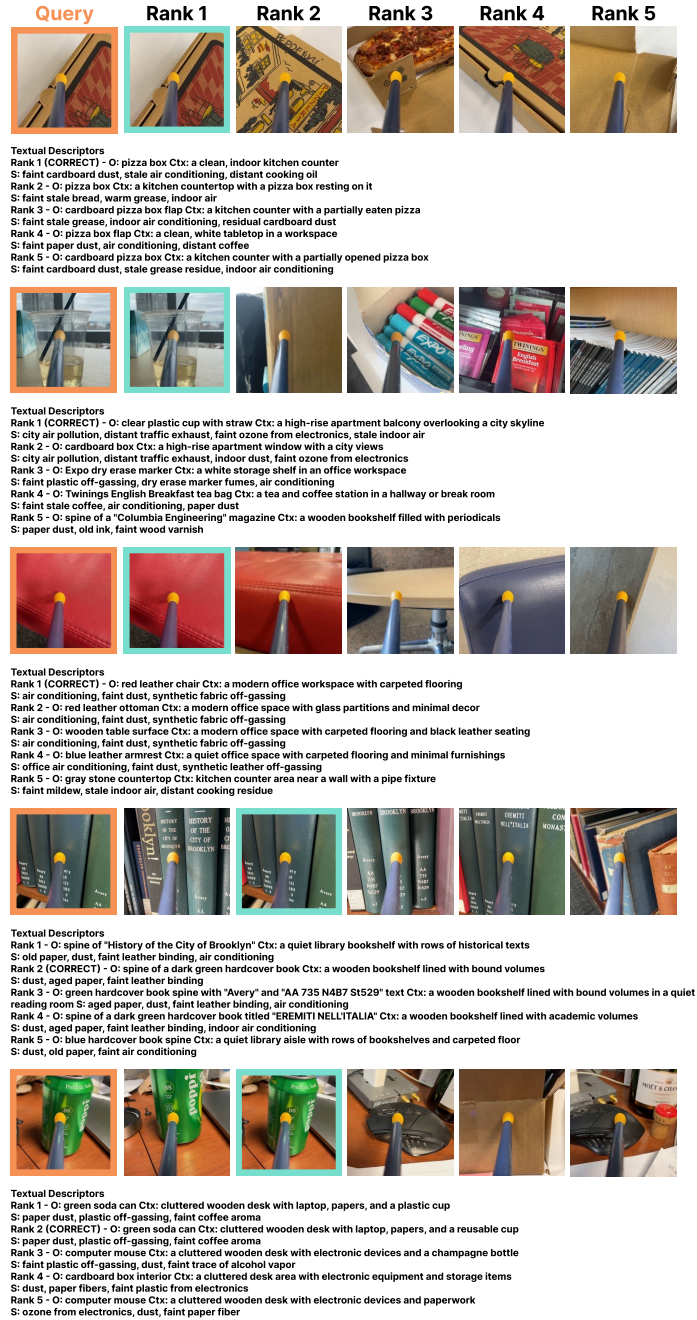}
\caption{\textbf{Extended S2IT Qualitative Results with Semantic Textual Descriptors.} We visualize the top-5 retrieved (Image, Text) pairs for a query olfactory signal (\textbf{\textcolor{orange}{orange}}). The \textbf{\textcolor{smell_cyan}{cyan}} border indicates the Rank-1 result. The accompanying text, decomposed into Object (\textbf{\textcolor{obj_color}{O}}), Context (\textbf{\textcolor{ctx_color}{Ctx}}), and Inferred Smells (\textbf{\textcolor{inf_color}{S}}), demonstrates that \mname{} aligns olfactory signals with complex scenes.}
\label{fig:qualitative_supp}
\end{figure}

We provide extended qualitative results for the Smell-to-Image-Text (S2IT) retrieval task in Figure~\ref{fig:qualitative_supp}. Each row displays the query olfactory signal alongside the top-5 retrieved (Image, Text) pairs.

\paragraph{Semantic Consistency and Error Analysis.}
In several instances (\eg, Row 1: Pizza, Row 4: Library Books), \mname{} achieves perfect or near-perfect retrieval, correctly identifying both the object and the environment. Notably, the textual descriptors for Row 4 reveal that the model distinguishes between specific attributes like \textit{"aged paper"} and \textit{"faint leather binding"}.

\paragraph{Ambient Environment Overlap.}
The textual descriptors explain many of the "near-misses." In Row 2 (Plastic Cup), while the Rank 2 result is a cardboard box, the text reveals a shared environmental context: \textit{"high-rise apartment balcony/window with a city view"} and shared ambient smells of \textit{"city air pollution"} and \textit{"traffic exhaust."} This confirms that even when the primary object differs, the model successfully identifies the broader olfactory scene, aligning the e-nose signal with the background atmosphere of the environment.

\clearpage


\end{document}